%% file: main.tex
\documentclass{article}
\usepackage{iclr2027_conference,times}
\input{math_commands.tex}

\usepackage{xurl,booktabs,array,graphicx,xcolor}
\usepackage{amsmath,amssymb}
\usepackage{flafter}
\usepackage{placeins}
\usepackage{subcaption}
\usepackage{algorithm}
\usepackage{algpseudocode}
\usepackage{hyperref}
\usepackage[capitalise,nameinlink,noabbrev]{cleveref}
\usepackage{giver_prompt_boxes}
\crefname{appendix}{Appendix}{Appendices}
\Crefname{appendix}{Appendix}{Appendices}
\definecolor{refblue}{HTML}{174A7E}
\hypersetup{colorlinks=true,linkcolor=refblue,citecolor=refblue,urlcolor=refblue,
  pdfauthor={Zinuo Zhang, Jian Yang, Xincan Wang, Yanchi Chen, Fanye Kong, Zhaoxin Fan, Xiaofeng Wang, Guan Huang, Borui Zhang, Zheng Zhu},
  pdftitle={Verify What Matters: Cost-Aware Long Video Generation with GIVER}}
\newcommand{\method}{GIVER}

\title{Verify What Matters:\\Cost-Aware Long Video Generation with GIVER}
\newcommand{\bylinename}[2]{\mbox{#1\textsuperscript{#2}}}
\newcommand{\bylinegap}{\hspace{0.72em}}
\newcommand{\bylinerow}[1]{\makebox[\linewidth][c]{#1}}
\newcommand{\affiliation}[2]{\mbox{\textsuperscript{#1}\,#2}}
\author{%
  \bylinerow{%
    \bylinename{Zinuo Zhang}{1,*}\bylinegap
    \bylinename{Jian Yang}{2,*}\bylinegap
    \bylinename{Xincan Wang}{3,*}\bylinegap
    \bylinename{Yanchi Chen}{1}\bylinegap
    \bylinename{Fanye Kong}{1}}\\
  \bylinerow{%
    \bylinename{Zhaoxin Fan}{4}\bylinegap
    \bylinename{Xiaofeng Wang}{1}\bylinegap
    \bylinename{Guan Huang}{5}\bylinegap
    \bylinename{Borui Zhang}{1,\textdagger}\bylinegap
    \bylinename{Zheng Zhu}{5,\textdagger}}%
}
\makeatletter
\renewcommand{\@maketitle}{%
  \par\begingroup
    \centering\setlength{\parskip}{0pt}%
    {\normalfont\bfseries\fontsize{18}{22}\selectfont \@title\par}
    \vspace{12pt}
    {\normalfont\fontsize{11.5}{16}\selectfont \@author\par}
    \vspace{6pt}
    {\normalfont\fontsize{10}{13.5}\selectfont
      \affiliation{1}{Tsinghua University}\qquad
      \affiliation{2}{Renmin University of China}\\
      \affiliation{3}{Shanghai Theatre Academy}\qquad
      \affiliation{4}{Beihang University}\qquad
      \affiliation{5}{GigaAI}\par}
  \endgroup
  \begin{NoHyper}
    \footnotetext[1]{\,Equal contribution.\qquad
      \textsuperscript{\textdagger}\,Co-corresponding authors.}
  \end{NoHyper}
  \vskip 13pt}
\makeatother
\iclrfinalcopy

\begin{document}
\maketitle
\fancyhead{}
\renewcommand{\headrulewidth}{0pt}

\begin{abstract}
Although recent agentic systems can generate long multi-shot videos from storyboards, physical inconsistencies can persist, such as fingers passing through a cup during lifting.
Repairing these defects requires identifying failures and guiding regeneration without altering the requested content.
To this end, we introduce \method{}, an agentic framework that embeds verification-and-repair loops into the generation process. 
Specifically, \method{} constructs a fixed Subject--Object Interaction Graph (SOIG), whose nodes encode storyboard-derived visual requirements and whose edges encode their logical prerequisites.
Using this graph, \method{} verifies reusable assets and repairs them when needed, then runs two local loops: L1 checks the first frame's initial state, while L2 checks required video behavior and visible artifacts, retaining the selected first frame.
Failed-node evidence guides targeted revisions to generation fields, and each new candidate is rechecked against the same requirements.
On ViMax-Bench and ST-Bench-10, \method{} leads the compared pipelines in fidelity and physical plausibility under automated and blinded human evaluation, with automated severe-artifact rates of 0.70\% and 2.80\%, respectively.
Additionally, our ablation study shows improved quality with approximately 24--25\% fewer video generation and editing operations than two flat-verification configurations.
\end{abstract}

\section{Introduction}
\label{sec:introduction}

A long video can be convincing for minutes and lose its credibility in a single shot. 
For example, when a character lifts a cup, his fingers may pass through cup's surface, undermining the physical plausibility of the whole video. 
While recent agentic systems~\citep{huang2025filmaster,huang2026vimax,wu2025movieagent} can produce long multi-shot videos via coordinate planning and synthesis, such defects still persist. 
A human editor can address these defects by inspecting the clip, identifying the failed interaction, and revising the prompt before regeneration, often over multiple attempts.
However, automating this diagnosis-and-repair process with agents remains challenging, especially when revisions must preserve the requested content.
The practical challenge is therefore \textbf{to reduce artifact-induced rework without simplifying the requested content or lowering its visual quality}.

\begin{figure}[t]
  \centering
  \vspace{-0.5cm}
  \includegraphics[width=\linewidth]{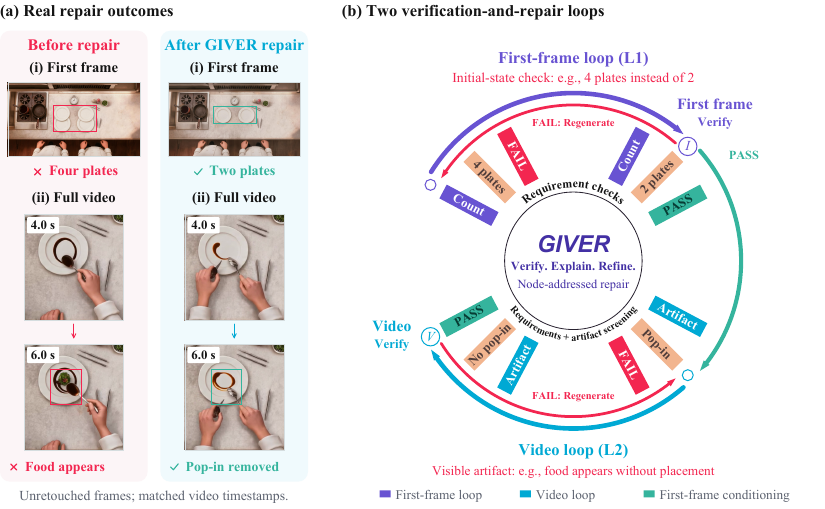}
  \caption{\textbf{Repair the initial state before repairing its evolution.} Real generated media show an initial-state mismatch (four plates instead of two) and a temporal artifact (unexplained food appearance). L1 and L2 have separate local regeneration paths; L2 retains the selected first frame. Asset verification precedes these loops. Source traces and evaluation protocols appear in \cref{app:figures}.}
  \label{fig:teaser}
\end{figure}

Automating this repair process requires a reliable link between visual diagnosis and the next generation request.
A global score can indicate that one shot video is better than another, but does not necessarily explain what went wrong~\citep{guan2025etva}.
By contrast, question-based evaluation, dependency-aware assessment, and localized visual feedback provide more specific diagnoses~\citep{hu2023tifa,cho2024dsg,liang2024richhf}; for example, checking hand--cup contact requires first establishing that both the hand and cup are present.
Yet identifying a failed check does not by itself specify how to revise the generation prompt.
If the fingers pass through the cup, the revision should address the contact while retaining the requested lifting action, rather than remove the action to avoid the defect.
Effective repair therefore requires keeping the original requirements fixed and using the observed failure to guide changes to the relevant parts of the prompt~\citep{lee2026videorepair}.

To address this limitation, we introduce \method{} (Graph-guided Iterative Verification and Evidence-linked Repair) around one principle: \emph{keep the requirements fixed, and use visual evidence to guide repairs for failed requirements}.
Specifically, we construct a Subject--Object Interaction Graph (SOIG) from the storyboard, where nodes encode atomic visual requirements and edges encode the logical prerequisites among them~\citep{cho2024dsg,pothiraj2026pqsg}.
Keeping this graph fixed makes individual failures traceable across attempts, while its dependencies prevent a check from receiving positive credit when its prerequisites have not been established.
Furthermore, to apply this representation during generation, \method{} draws on the filmmaking workflow of preparing a scene before filming it, embedding verification-and-repair loops into both first-frame and video generation.
Accordingly, \method{} first verifies and repairs reusable assets, then runs two local verification-and-repair loops: L1 checks the first frame's initial state; L2 checks required behavior and visible video artifacts (\cref{fig:teaser})
In both loops, failed checks and their visual evidence guide revisions to the relevant generation fields.
Each new candidate is then checked against all fixed requirements to determine whether the repair resolved the targeted failure without introducing new errors.

On ViMax-Bench~\cite{huang2026vimax} and ST-Bench-10~\cite{sun2025t2vcompbench}, \method{} achieves higher automated fidelity and physical-plausibility scores and lower severe-artifact rates than the compared pipeline adaptations. Blinded human ratings also favor \method{} on fidelity and physical plausibility. In a separate video-stage experiment with shared baseline-generated first frames, \method{} improves quality with approximately 24--25\% fewer video generation and editing operations than two flat-verification configurations. 
\textbf{Our~contributions~are:} \textbf{(i)} a dependency-aware graph linking fixed requirements and node-level visual evidence to targeted repair; \textbf{(ii)} a staged pipeline combining asset verification and repair with separate first-frame and video loops; and \textbf{(iii)} automated and blinded human evaluations of video quality, plus a fixed-first-frame experiment on repair efficiency.

\section{Related Work}
\label{sec:related}

Multi-shot video systems must coordinate a narrative and control the quality of its realization. We distinguish three control mechanisms by the decision they make: \emph{what to generate}, \emph{which candidate to retain}, and \emph{what to change after observing an output}. A system can combine these mechanisms; the distinction concerns their control decisions, not agent names or a ranking of architectures.

\paragraph{Planning and conditioning.}
This family controls the information supplied to generation. Vlogger, VideoDirectorGPT, and MovieAgent decompose narratives into executable scene or shot plans~\citep{zhuang2024vlogger,lin2024videodirectorgpt,wu2025movieagent}. VideoStudio, VGoT, and StoryAgent carry entity references or visual context across shots~\citep{long2024videostudio,zheng2024vgot,hu2024storyagent}; Captain Cinema and STAGE use keyframes or structured storyboards to coordinate synthesis~\citep{xiao2025captaincinema,zhang2025stage}. Related model-level approaches include hierarchical sequence generation in MovieDreamer and joint multi-shot modeling in HoloCine~\citep{zhao2025moviedreamer,meng2025holocine}. StoryMem and OneStory maintain visual memory for later shots~\citep{zhang2025storymem,an2025onestory}. Their shared purpose is to make independently generated content consistent with a larger narrative. Such conditioning is complementary to checking whether the resulting images and clips correctly depict the requested objects, local interactions, and motion.

\paragraph{Evaluation-guided selection.}
Here, evaluation determines which candidate to retain or where to search next. Anim-Director evaluates visual candidates, and ViMax uses VLM-based best-of-$k$ selection~\citep{li2024animdirector,huang2026vimax}. AniMaker extends this mechanism to contextual clip evaluation with MCTS-driven search~\citep{shi2025animaker}. Their common decision concerns \emph{which realization to use}. Selection can improve quality without specifying a corresponding change to a generation field. ViMax also uses a graph to coordinate visual references across shots; \method{} instead represents prerequisites among checks within a shot. It retains candidates and links their failed checks to generation fields, using the diagnosis to direct the next repair request.

\paragraph{Feedback-driven correction.}
This mechanism uses generated outputs to revise later requests, intermediate content, or the workflow itself. DreamFactory iterates keyframes, while MAViS and PersonaVlog integrate staged review and refinement~\citep{xie2024dreamfactory,wang2025mavis,hou2025personavlog}. AesopAgent and VideoWeaver improve workflows or reusable skills, and Hollywood Town allows bounded feedback between agents~\citep{wang2024aesopagent,wei2026videoweaver,wei2025hollywoodtown}. FilMaster uses audience-oriented feedback in post-production~\citep{huang2025filmaster}; Co-Director combines local refinement with final-video rewards for creative-direction search~\citep{song2026codirector}. \method{} belongs to this family. Its contribution is not the existence of a feedback loop, but the fixed requirements and node-linked observations that connect a failed candidate to an eligible revision field. These requirements persist across attempts, so each revision is evaluated against the same target.

\paragraph{Structured visual feedback.}
Question-based evaluation decomposes alignment into atomic checks. TIFA evaluates image faithfulness, DSG introduces question dependencies, and ETVA and PQSG extend structured assessment to video alignment and physics~\citep{hu2023tifa,cho2024dsg,guan2025etva,pothiraj2026pqsg}. NeuS-V uses temporal logic, while SG-PVR grounds video reward reasoning in scene graphs~\citep{sharan2025neusv,kim2026sgpvr}. These are evaluation representations rather than complete long-video production pipelines. Related generation methods use physics-oriented prompt refinement, multi-agent critique, or localized correction~\citep{xue2025phyt2v,long2026vista,lee2026videorepair}. Particularly close, \citet{tyagi2026beyond} combine alignment questions and artifact checks with prompt and parameter optimization for image-to-video generation. \method{} builds on these foundations through fixed graphs that connect the same requirements to fresh visual evidence across repair attempts, with checks for assets, first frames, and videos.

\section{Method}
\label{sec:method}

\begin{figure}[t]
  \centering
  \includegraphics[width=\linewidth]{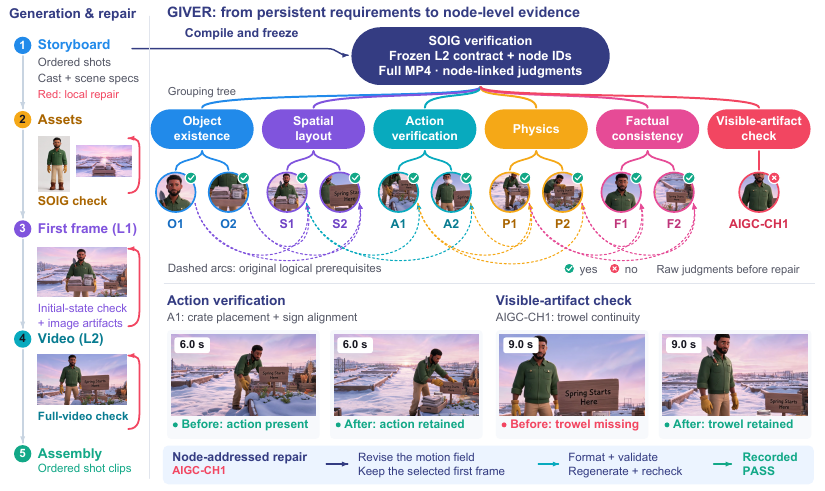}
  \caption{\textbf{From a storyboard to evidence-linked repair.} Assets, first frames, and videos undergo local review before ordered assembly. The expanded L2 tree groups atomic checks; dashed links represent logical prerequisites. Real frames show a satisfied action check (A1) and a failed trowel-continuity guard (AIGC-CH1), which guides motion-field revision while retaining the first frame. Badges denote recorded verifier judgments; \cref{app:figures} gives the recorded trace provenance.}
  \vspace{-0.5cm}
  \label{fig:pipeline}
\end{figure}

\method{} converts storyboard requirements into a graph, checks each candidate against those requirements, and revises the generation fields associated with failed checks. \Cref{fig:pipeline} shows the procedure, from reusable assets to clip assembly. We first define the task and the fixed requirement graph, then explain how node-level evidence determines acceptance and guides the controller's next repair.

\subsection{Task and Objective}
\label{sec:problem}

The input is a storyboard $X=(x_1,\ldots,x_N)$ with cast and scene specifications. Each shot $x_i$ describes an initial state, action, camera, and duration $d_i$. An upstream planner may produce this storyboard; our control procedure begins with it. Let $G_I$ and $G_V$ denote black-box image and image-conditioned video generators. Given selected references $\mathcal A_i$, image and video attempts follow
\begin{equation}
 I_i^{(a)}\sim G_I(p_{i,I}^{(a)},\mathcal A_i),\qquad
 v_i^{(b)}\sim G_V(p_{i,V}^{(b)},I_i^*;d_i),\qquad
 y=\operatorname{Concat}(v_1^*,\ldots,v_N^*).
 \label{eq:generation}
\end{equation}
Here $a,b$ index attempts, $p$ is a generation request, and $I_i^*,v_i^*$ are the selected frame and clip. L2 holds $I_i^*$ fixed. The procedure uses only generator requests, without access to weights or gradients.

The objective is fewer visible artifacts and fewer regeneration attempts while preserving fidelity to $X$. An artifact is a failure of visual integrity or physical plausibility, not simply a missing scripted action. We therefore assess fidelity, artifact severity, and effort separately. Stage budgets limit generation, verification, and refinement requests. A playable output selected at budget exhaustion is distinguished from an output that passes verification. This distinction separates the decision to stop spending requests from the evidence that a selected candidate meets the visual requirements.

\subsection{A Fixed Graph of Visual Requirements}
\label{sec:contracts}

For shot $i$ and stage $\ell\in\{I,V\}$, the compiler builds a directed acyclic graph $\mathcal C_{i,\ell}=(V_{i,\ell},E_{i,\ell})$. We call this set of requirements a \emph{contract}. The image graph describes the initial state; the video graph adds actions and temporal requirements. Asset contracts are compiled separately. Each node contains an identifier, question, target condition, relevant entities, category, criticality, and evidence requirement. Categories cover object existence ($O$), spatial layout ($S$), action ($A$), physics ($P$), and factual consistency ($F$). An edge records a logical prerequisite: checking hand--cup contact requires identifying both the hand and cup. It does not describe a transition between shots. Nor does a category-grouping branch in \cref{fig:pipeline} imply a prerequisite: the dashed links encode those dependencies. The compiler validates node identities, edge endpoints, acyclicity, and stage compatibility, so that a still image is checked for its initial state, not for evidence of actions that unfold over time.

The graph remains fixed across attempts within its stage. Stages share the representation and storyboard, not an identical set of nodes. Runtime rules specify applicable checks, artifact guards, and any exclusions or waivers; these decisions are recorded separately from visual observations. This preserves three distinct objects: what the shot requires, what the verifier observes, and how the controller decides to proceed. A waiver can affect the controller's acceptance decision, but an observed failure must remain recorded as a failure rather than be presented as evidence of success.

\subsection{Verification with Evidence and Dependencies}
\label{sec:verification}

For a candidate, the verifier returns a record for each requested node $j$:
\begin{equation}
 r_j=(a_j,o_j,z_j),\qquad a_j\in\{\mathrm{yes},\mathrm{no},\mathrm{unknown}\},
 \label{eq:record}
\end{equation}
where $o_j$ describes the observation and $z_j$ specifies its supporting evidence. Each record also identifies the shot, stage, candidate, and node, so observations can be traced to the evaluated media. An unknown judgment denotes insufficient evidence, not an observed defect or a confirmed success.

L1 uses multi-turn requirement review with image reinspection and an artifact check. In the current L2 implementation, a separate conversation for each candidate receives the complete MP4 and queries the fixed nodes in dependency order, one node per turn. Earlier validated answers remain in context but cannot be overwritten. Video answers must refer to the clip: an optional first-frame attachment cannot establish later behavior. Artifact guards use the same interface. Checks on the response format and node coverage reject missing or mismatched identities. The resulting record identifies the evaluated requirement, links the answer to its supporting visual evidence, and preserves unresolved judgments for the controller instead of forcing a definite positive or negative verdict.

Let $V_s$ be the active scoring nodes, $b_j$ their policy-adjusted satisfaction, and $\operatorname{Anc}(j)$ the active ancestors of $j$. A failed prerequisite blocks positive credit below it:
\begin{equation}
 \widetilde b_j=b_j\prod_{u\in\operatorname{Anc}(j)}b_u,\qquad
 s_{\mathrm{tree}}=\frac{1}{|V_s|}\sum_{j\in V_s}\widetilde b_j,\qquad
 s_{\mathrm{base}}=1+9s_{\mathrm{tree}}.
 \label{eq:score}
\end{equation}
The empty-graph tree score is zero. Critical failures and artifact/physics vetoes additionally affect acceptance. Unknown evidence does not count as observed satisfaction; policy adjustments are logged separately. This diagnostic score controls generation and is not an external evaluation metric. Consider a contact node whose two object-presence prerequisites are satisfied: its own evidence must still establish contact. Conversely, if a prerequisite is unresolved, a downstream affirmative answer cannot independently establish the interaction. The raw answers distinguish an observed contact failure from insufficient evidence about an object. The graph determines how node answers support a decision while preserving the original observations for inspection and later repair.

\subsection{Local Repair and Candidate Selection}
\label{sec:repair}
\label{sec:stages}
 
\begin{algorithm}[H]
 \caption{Local verification and repair in \method{}}
 \label{alg:repair}
 \begin{algorithmic}[1]
  \Require Fixed stage graph $\mathcal C$, initial request $p$, conditioning $c$, request limits $B$
  \State $\mathcal H\gets\emptyset$ \Comment{Candidate records, including failed and unknown checks}
  \While{the stage request limits permit another attempt}
   \State Generate or edit a candidate $m$ using $(p,c)$; account for requests in $B$
   \State Validate media and obtain node observations against $\mathcal C$ within $B$
   \If{media or report validation fails}
    \State Record the failure; apply configured bounded error handling
    \State \textbf{continue} if permitted; otherwise \textbf{break}
   \EndIf
   \State Compute the dependency-aware score and acceptance decision
   \State Append media, raw observations, policy decisions, and request to $\mathcal H$
   \State \textbf{if} accepted \textbf{then return} candidate and its recorded report
   \State Identify repair targets and eligible fields from failed or unresolved checks
   \State \textbf{if} no eligible repair or insufficient requests remain \textbf{then break}
   \State Revise and validate complete text for eligible fields; retain other fields
   \State Keep $\mathcal C$ and stage conditioning $c$ fixed for the next candidate
  \EndWhile
  \State Apply the configured fallback policy to $\mathcal H$, or report no selected output
  \State \Return selected media, if any, with its original report and termination reason
 \end{algorithmic}
\end{algorithm}

The controller applies verification in production order. It first reviews reusable character and static scene/object assets with identity and image checks. Failed checks trigger bounded asset repair before the selected references are attached to downstream requests. L1 then generates a first frame and checks its initial state and image artifacts. For example, four plates instead of two calls for a change to the first-frame description. An accepted frame, or a recorded fallback at exhaustion, becomes the video condition. L2 checks the resulting clip for required behavior and visible artifacts, including temporal failures that a still image cannot reveal. \textbf{L2 does not return to L1}: video repair keeps the selected frame fixed. Selected clips are finally assembled in storyboard order for final delivery.

Within each local loop, failed or explicitly unresolved nodes identify eligible revision fields. The refiner receives their observations, candidate media, and current descriptions, then returns the \emph{complete revised text} of each affected field. Other fields retain their values. L1 revises the image description; L2 revises video-generation fields without replacing its first frame. Before submission, the controller validates the update, reference attachments, and provider formatting. The graph is never rewritten to excuse a failure. A text revision can still affect unrelated pixels, so each new candidate is checked against all fixed requirements, including requirements that passed in earlier generation attempts.

\Cref{alg:repair} summarizes this shared local procedure. Candidate records link the report, revised fields, parent candidate, and submitted request. Limits cover media generation, verification, refinement, response corrections, and retries, rather than only the outer loop count. An enabled fallback policy may select existing valid, scored media at exhaustion, but preserves its failed and unknown checks. Delivery is not relabeled as acceptance. \Cref{app:implementation} specifies execution details and the run-specific verification protocols; the video-stage experiment evaluates the graph-based repair configuration, not the isolated effect of the current one-node-per-turn interface used to query the video.

\section{Experiments}
\label{sec:experiments}

We test three claims: \method{} reduces visible artifacts while preserving requested content; the quality gains are also apparent to human viewers; and structured video verification reduces repair operations relative to flat verification. Automated system comparisons, blinded human ratings, and a fixed-first-frame control experiment address these claims in turn. As in VBench, EvalCrafter, and T2V-CompBench~\citep{huang2024vbench,liu2024evalcrafter,sun2025t2vcompbench}, we report separate scores for fidelity to requested content and for physical plausibility rather than combining both into one score.

\subsection{Benchmarks and Evaluation Setup}
\label{sec:setup}

\paragraph{Tasks and generators.}
ViMax-Bench~\citep{huang2026vimax} provides 35 themes: 12 single-character, 12 multi-character, and 11 multi-scene cases. We use its supplied scripts. Copyright restrictions limit our ST-Bench~\citep{zhang2025storymem} evaluation to ten cases, which we denote \emph{ST-Bench-10}. ViMax, MovieAgent, AniMaker, and \method{} are run on the same ten cases, and their generated outputs are scored by the MLLM evaluator; \cref{tab:automatic} reports the results. ST-Bench-10 uses five-second clips. The rendering stack uses Seedream 5.0 and Seedance 2.0, 720p, 16:9, with a shared stylized 3D setting. The ViMax system baseline uses 15-second clips, whereas the separate video-stage ablation uses ten-second clips and supplies the same baseline first frame to the repair configurations.

\paragraph{Baseline configurations.}
We adapt ViMax, MovieAgent, and AniMaker to the same rendering backend. {AniMaker retains its MCTS-based search framework with next-shot lookahead disabled ($w_1=3,w_2=0$; \cref{app:baseline-configurations}).} \Cref{tab:automatic} compares the outputs from these configurations. The human study uses separately sampled, blinded four-system groups (\cref{sec:human}). \Cref{app:evaluation} records the benchmark cases and aggregation rules for each experiment.

\paragraph{Post-generation assessment.}
ShotEval-MLLM uses Doubao Seed 2.0 Pro with script-derived checks for fidelity, artifacts, and physics. Frame sampling combines 16 overview, 48 motion, and eight peak frames, with eight-fps local inspection. Artifact and global-commonsense judgments are prompt-blind; script and rule checks use the reference requirements. Assessment is run separately from generation but shares the internal verifier's model family. We therefore complement it with blinded human ratings rather than treating either model score as an independent perceptual ground truth~\citep{liu2025reliablejudge}. \Cref{app:evaluation} specifies metric denominators and exclusions.

\subsection{Metrics and Automated Results}
\label{sec:automatic}

\textbf{SF} is script fidelity: passed atomic claims divided by all passed, failed, and uncertain claims. \textbf{Phys} is the 1--5 global physical-commonsense score; the ViMax-Bench aggregate conditions on physics-evaluable shots. \textbf{Severe\%} is the fraction of evaluated shots with at least one confirmed artifact of severity 3, denoting an obvious generation failure. \textbf{Joint} counts physics-evaluable shots scoring at least 4, without an additional SF threshold. Its \emph{denominator differs}: physics-evaluable shots on ViMax-Bench, but all evaluated shots on ST-Bench-10. Joint values must therefore be compared within each benchmark block, since the two denominators measure different proportions.

\begin{table}[!htbp]
  \centering
  \small
  \caption{\textbf{Automated evaluation.} \method{} leads all four reported metrics on both benchmarks. Bold marks the best result. Results summarize available outputs from the adapted pipelines in \cref{sec:setup}; Joint has benchmark-specific denominators and is compared within each benchmark block.}
  \label{tab:automatic}
  \begin{tabular}{@{}llrrrr@{}}
    \toprule
    Benchmark & System & SF $\uparrow$ & Phys $\uparrow$ & Joint $\uparrow$ & Severe\% $\downarrow$ \\
    \midrule
    ViMax-Bench & ViMax & .769 & 3.760 & .512 & 3.50 \\
     & MovieAgent & .753 & 3.790 & .531 & 4.20 \\
     & AniMaker & .768 & 3.650 & .515 & 5.70 \\
     & \method{} & \textbf{.780} & \textbf{4.010} & \textbf{.627} & \textbf{0.70} \\
    \midrule
    ST-Bench-10 & ViMax & .687 & 3.944 & .278 & 9.70 \\
     & MovieAgent & .700 & 4.278 & .361 & 5.60 \\
     & AniMaker & .738 & 4.097 & .306 & 6.90 \\
     & \method{} & \textbf{.791} & \textbf{4.361} & \textbf{.472} & \textbf{2.80} \\
    \bottomrule
  \end{tabular}
\end{table}

\paragraph{Fewer artifacts without lower fidelity.}
\method{} has the highest SF, Phys, and within-benchmark Joint, and the lowest Severe\% in both blocks of \cref{tab:automatic}. On ViMax-Bench, its severe-artifact rate is 0.70\%, compared with 3.50--5.70\% for the adapted baselines; SF is .780 versus .753--.769. ST-Bench-10 shows the same direction: 2.80\% severe artifacts and .791 SF, versus 5.60--9.70\% and .687--.738. Relative to the strongest baseline on Severe\%, these are reductions of 80\% and 50\%, respectively. Higher fidelity accompanies the artifact reduction, so the improvement is not explained by a lower aggregate rate of satisfying the requested content. ViMax-Bench averages at case level, whereas ST-Bench-10 pools evaluated shots; comparisons are made within each benchmark.

\begin{figure}[t]
  \centering
  \begin{subfigure}[t]{0.5409\linewidth}
    \caption{Woodworking: geometry and contact}
    \label{fig:qualitative-woodworking}
    \includegraphics[width=\linewidth]{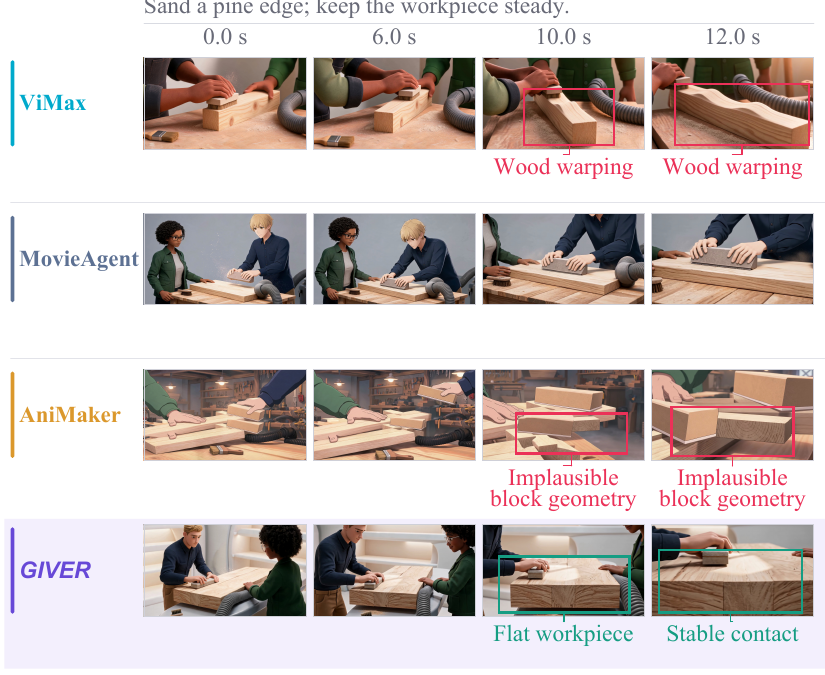}
  \end{subfigure}%
  \begin{subfigure}[t]{0.4591\linewidth}
    \caption{Wine counter: temporal continuity}
    \label{fig:qualitative-wine}
    \includegraphics[width=\linewidth]{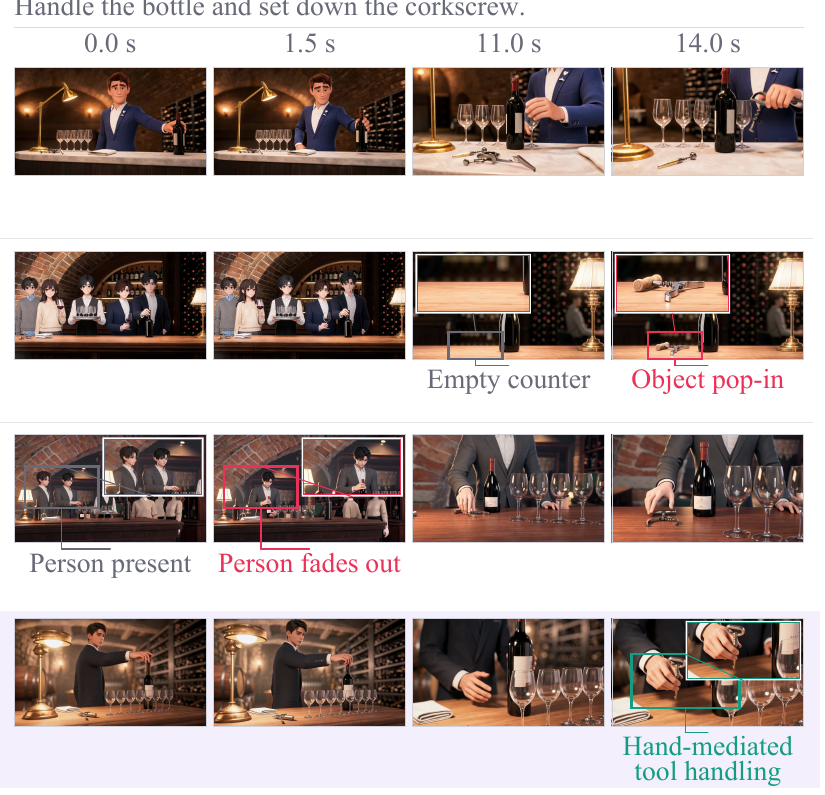}
  \end{subfigure}
  \caption{\textbf{Qualitative comparison on two ViMax-Bench examples.}
  Each row shows one pipeline, with four matched-time frames per example.
  \textbf{(a)} ViMax develops a warped workpiece and AniMaker shows implausible block geometry.
  \textbf{(b)} AniMaker's background figure fades without leaving the view; MovieAgent's corkscrew appears without visible placement.
  \method{} shows a flat workpiece and hand-mediated tool handling in the illustrated sequences.
  Boxes locate the observations, and insets enlarge the same frames. These are selected examples, not estimates of failure frequency.}
  \vspace{-0.5cm}
  \label{fig:qualitative-comparison}
\end{figure}

\paragraph{Qualitative comparison.}
\Cref{fig:qualitative-comparison} compares four methods at matched timestamps on two ViMax-Bench tasks. Woodworking exposes geometric distortion during contact; the wine-counter task exposes unexplained temporal changes. Labels identify observed defects, not a global judgment on every frame. \method{} preserves a flat workpiece and shows hand-mediated tool handling in the displayed sequences. Both panels retain all four methods to make the comparison explicit. The examples also explain the need for two kinds of evidence: object shape can be inspected in a frame, whereas unexplained appearance requires comparing what was visible before and after. These examples retain the requested objects and actions while illustrating their visual integrity over time.

\subsection{Blinded Human Evaluation}
\label{sec:human}

Raters assessed script fidelity on a 0--1 scale and physical plausibility on a 1--5 scale. Systems were concealed behind neutral codenames, and shot order and within-shot system order were randomized. For ViMax-Bench, 16 raters each assessed 12 sampled shot groups with four systems, producing 768 valid ratings. The combined sample contains 37 shots and 148 system--shot videos. For ST-Bench-10, 25 raters assessed 24 shot groups, each containing one video per system, for 96 videos across the four systems. Of 2,400 submitted ratings, 78 had at least one out-of-range score and were excluded, leaving 2,322 valid ratings. \Cref{tab:human} averages valid ratings within each system. The human study uses a separately sampled set, distinct from the automated evaluation set.

\begin{table}[!htbp]
  \centering
  \small
  \caption{\textbf{Blinded human evaluation.} \method{} receives the highest mean SF and Phys on both samples. Scales are 0--1 and 1--5, respectively. Counts are valid rating records, not independent videos; bold marks the best mean for each metric within each benchmark's human sample.}
  \label{tab:human}
  \begin{tabular}{@{}lrrrcrrr@{}}
    \toprule
     & \multicolumn{3}{c}{ViMax-Bench} && \multicolumn{3}{c}{ST-Bench-10} \\
    \cmidrule(lr){2-4}\cmidrule(l){6-8}
    System & SF $\uparrow$ & Phys $\uparrow$ & Ratings && SF $\uparrow$ & Phys $\uparrow$ & Ratings \\
    \midrule
    ViMax & .577 & 3.266 & 192 && .526 & 3.625 & 579 \\
    MovieAgent & .562 & 3.000 & 192 && .543 & 3.412 & 578 \\
    AniMaker & .563 & 3.141 & 192 && .570 & 3.292 & 582 \\
    \method{} & \textbf{.629} & \textbf{3.599} & 192 && \textbf{.657} & \textbf{3.731} & 583 \\
    \bottomrule
  \end{tabular}
\end{table}

\paragraph{Viewers also rate content and plausibility higher.}
\method{} has the highest mean in both dimensions on both samples. Relative to the best baseline mean for each dimension, its SF margin is .052 on ViMax-Bench and .087 on ST-Bench-10; the corresponding physics margins are .333 and .106. These ratings support the quality advantage without exposing raters to the system identity or internal acceptance score. Agreement in the best-ranked system does not imply agreement on every individual video. \Cref{app:human} reports video-level correlations and the handling of repeated ratings; our conclusions concern the observed means, without a statistical-significance claim.

\subsection{Ablation: Video-Stage Verification}
\label{sec:ablation}

We study video-stage control in a ten-second generation experiment across all 35 ViMax-Bench themes. \method{} and both flat-verification variants use the baseline-generated first frame, with no new image-generation step. The comparison includes generation without SOIG verification, two event-list configurations (\texttt{abl1} and \texttt{abl2}), and \method{}'s graph-based video verification and repair. Both flat variants retain the baseline frame; their run identifiers do not denote different first-frame strategies in this experiment. \Cref{tab:ablation} includes 128 Type C \method{} clips from the separately implemented Type C pipeline. \Cref{app:versions} records the source of these Type C results, the retained evaluation samples, and the generation and verification protocols used by each configuration.

\begin{table}[h]
  \centering
  \small
  \caption{\textbf{Video-stage verification with fixed first-frame conditioning.} \method{} improves all four quality metrics with fewer video operations than either flat configuration. Video Try sums generation and editing per selected shot. All settings cover 35 themes at ten seconds per clip; valid-report counts and Type C provenance are given in \cref{app:versions}, along with run-specific protocols.}
  \label{tab:ablation}
  \begin{tabular}{@{}lrrrr@{}}
    \toprule
    Metric & Generation only & Flat (\texttt{abl1}) & Flat (\texttt{abl2}) & \method{} \\
    \midrule
    SF $\uparrow$ & .750 & .759 & .755 & \textbf{.766} \\
    Physics $\uparrow$ & 3.790 & 3.757 & 3.811 & \textbf{3.932} \\
    Joint $\uparrow$ & .566 & .552 & .560 & \textbf{.596} \\
    Severe\% $\downarrow$ & 5.5 & 4.8 & 5.5 & \textbf{3.6} \\
    \midrule
    Video Gen / Shot & 1.000 & 1.745 & 1.736 & 1.574 \\
    Video Edit / Shot & --- & 1.111 & 1.174 & .609 \\
    Video Try / Shot & 1.00 & 2.86 & 2.91 & 2.18 \\
    \bottomrule
  \end{tabular}
\end{table}

\paragraph{Quality and effort measures.}
SF and artifact assessment use the full script; the physics contract excludes camera motion. SF and Physics pool shot scores. Joint is the fraction of physics-evaluable shots scoring at least 4, and Severe\% averages theme-level rates. \Cref{app:versions} reports retained evaluation counts and run-specific verification protocols for each video-stage configuration.

\paragraph{Better quality with fewer video operations.}
\method{} achieves .766 SF, 3.932 Physics, and .596 Joint, with a 3.6\% severe-artifact rate. Both generation and editing decrease relative to flat verification, totaling 2.18 rather than 2.86 or 2.91 operations per selected shot: approximately 24--25\% fewer operations. Generation only uses one operation but has lower quality. The shared first frame rules out better image synthesis as the source of these gains. This compares complete video-stage configurations, not individual graph edges. \Cref{app:accounting} distinguishes video-operation counts from total costs including verification, prompt refinement, and other upstream model calls.

\begin{figure}[h]
  \centering
  \begin{subfigure}[t]{.49\linewidth}
    \caption{Script fidelity (SF $\uparrow$)}
    \includegraphics[width=\linewidth]{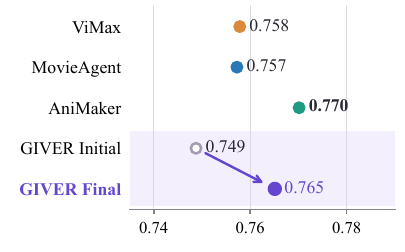}
  \end{subfigure}\hfill
  \begin{subfigure}[t]{.49\linewidth}
    \caption{Physical plausibility (Physics $\uparrow$)}
    \includegraphics[width=\linewidth]{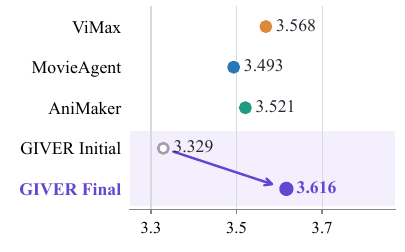}
  \end{subfigure}
  \vspace{-0.5cm}
  \caption{\textbf{Initial and final outputs on retried shots.} Means over 146 matched fifteen-second shots from 33 ViMax-Bench cases. Arrows connect the initial and final \method{} outputs. Baselines use the same shots. Historical runs were selected by case-level Physics (\cref{app:retry-analysis}).}
  \label{fig:retry-analysis}
\end{figure}

\paragraph{Initial-to-final quality on retried shots.}
We separately examine fifteen-second shots with multiple video attempts. Across 177 paired shots from 34 cases, mean Physics rises from 3.373 to 3.678 and SF from .750 to .762. On the 146 shots also scored for all baselines (\cref{fig:retry-analysis}), Physics rises from 3.329 to 3.616 and SF from .749 to .765. Final Physics is the highest mean on this subset; AniMaker has the highest SF. Each pair is evaluated against the original benchmark script, so both outputs are assessed for the same requested content. These gains describe selected historical runs, not uniformly beneficial retries; \cref{app:retry-analysis} details run selection and paired score changes.


\section{Discussion and Limitations}
\label{sec:limitations}

The results support a practical design choice: retain the task requirements and use specific evidence to guide repair. The graph makes decisions inspectable, but perception still depends on the verifier's temporal coverage, an issue also observed in video-understanding evaluation~\citep{nagrani2025minerva,fu2025videomme}. Keeping the first frame fixed saves repeated image generation but prevents L2 from revisiting an error missed by L1. Cross-shot requirements are a natural extension beyond the present shot-level contracts. Our experiments establish gains in measured clip quality and video-operation counts. {Graph compilation and verification average only \$0.1591/shot (7.12\% of one Seedance generation; \cref{app:verification-cost}). Full-story continuity and human inspection time merit further study.}

\section{Conclusion}

\method{} connects fixed, dependency-linked requirements to visual evidence and targeted repair for multi-shot video generation. Asset verification and repair precede separate first-frame and video loops; video repair retains the selected first frame. Automated and blinded human evaluations rank \method{} first among the compared pipeline adaptations in fidelity and physical plausibility. The fixed-first-frame experiment shows lower severe-artifact rates with about 24--25\% fewer video generation and editing operations than two flat-verification configurations. Actionable verification identifies failed checks, guides repair with evidence, and rechecks candidates against fixed requirements.

\clearpage
\subsection*{Ethics Statement}
Long-video generation can enable deceptive synthetic media, unauthorized likeness use, and biased portrayals. Our checks address visible defects and requested content, not consent, authenticity, or comprehensive content safety. The evaluation uses generated media and reports human ratings in aggregate; individual rater identities are not included in the manuscript. Automated judges and human ratings can reflect cultural and perceptual biases. Generated examples should be identified as synthetic media and should not be presented as authentic recordings or evidence of real events.

\subsection*{Reproducibility Statement}
\Crefrange{app:implementation}{app:figures} document the control interface, model identifiers, run-specific protocols, metric denominators, human-study aggregation, and figure provenance. The video-stage experiment specifies its fixed conditioning, Type C source, and media-operation accounting separately from the main system comparison. These details support reproduction of the reported evaluation and interpretation of each experimental setting; exact media regeneration remains subject to proprietary API behavior.

\subsection*{AI Use Statement}
AI assistants supported literature retrieval and synthesis, methodological discussion, code assistance, manuscript drafting and editing, and figure-production assistance. Generative models also produce and evaluate the experimental media as described in the method and evaluation sections. The authors are responsible for the scientific claims, references, data interpretation, and final manuscript.

\bibliography{references}
\bibliographystyle{iclr2027_conference}
\clearpage
\appendix
\crefalias{section}{appendix}
\crefalias{subsection}{appendix}
\input{appendix}
\input{appendix_mllm_prompts}
\input{appendix_rebuttal}
\input{appendix_refiner_prompts}
\end{document}

%% file: math_commands.tex
\usepackage{amsmath,amsfonts,bm}

\def\eqref#1{equation~\ref{#1}}

\def\1{\bm{1}}

\DeclareMathAlphabet{\mathsfit}{\encodingdefault}{\sfdefault}{m}{sl}
\SetMathAlphabet{\mathsfit}{bold}{\encodingdefault}{\sfdefault}{bx}{n}



%% file: appendix.tex
\section{Control and Implementation Details}
\label{app:implementation}

\subsection{Generation Procedure}
The independently generated first-frame route proceeds as follows.
\begin{enumerate}
 \item Validate the storyboard, generator interfaces, reference attachments, and stage budgets. Prepare reusable assets; apply identity-focused or static-image checks and bounded repair.
 \item Compile separate first-frame and video contracts from the shot and asset metadata. Validate graph identities and logical prerequisites. Record which checks apply at runtime and keep policy adjustments separate from the fixed requirements and raw visual observations.
 \item Generate a first-frame candidate, perform L1 requirement review and image-artifact screening, and retain the observations and the controller's recorded acceptance decision.
 \item On a repairable non-pass, revise eligible image fields and generate or edit another candidate. Apply only configured, bounded same-image rechecks for unresolved evidence. Select a passing frame, or an explicitly permitted scored fallback at resource exhaustion.
 \item Bind the frame; check formatting and repair application before video submission.
 \item Verify the resulting clip, record node evidence and the acceptance decision, and revise video fields when a repair target and budget remain. The selected frame remains fixed.
 \item Select the video under the recorded acceptance or exhaustion policy. Preserve any failed or unknown checks. Assemble selected clips in storyboard order and report which shots produced an output, separately from whether those outputs passed their quality checks.
\end{enumerate}

\subsection{Verification, Policy, and Candidate Records}
L1's second requirement-review turn reinspects the image with provisional observations; it is not merely an answer-formatting step. Image artifacts and portrait identity use their applicable checks. The current L2 conversation maintains a frozen node order and candidate-specific media context. Each answer identifies its node and references the video. Missing, duplicate, or mismatched identifiers invoke bounded response handling rather than automatic acceptance. Earlier validated answers remain in context and cannot be revised by a later response in the same candidate session.

Full-video attachment is not exhaustive observation of all frames. Perception depends on provider preprocessing and sampling, and prerequisite answers can bias later judgments. Artifact guards distinguish insufficient visibility and authored occlusion from confirmed unexplained disappearance. Node-level evidence is therefore a diagnostic record, not a formal proof of correctness.

The base graph, runtime applicability, raw answers, policy adjustments, and gate outcome are distinct objects. Critical failures or artifact/physics vetoes can prevent acceptance despite a high tree score. An unresolved or dependency-blocked node's zero score contribution is not itself a confirmed artifact. A policy waiver does not retrospectively change a raw observation to \texttt{yes}.

The refiner returns complete revised text for eligible fields, not an unconstrained replacement storyboard. Validation checks node references, field eligibility, usable text, and an effective revision. Formatting the revised request for a provider is a separate operation. Fields outside the update scope remain unchanged, but semantics inside a rewritten field and the resulting pixels may change.

Only valid media with a usable report can enter scored-candidate retention. A fallback records the original verdict, failed nodes, uncertainty, and termination reason. A parser failure, invalid media file, or missing repair target is not permission to relabel a candidate as passed. Accounting distinguishes physical generation, editing, verification, refinement, and failed requests; overlapping service calls must be accounted for separately rather than summed to estimate wall-clock latency.

\subsection{Run Configurations and Protocol Versions}
\label{app:versions}

The model identifiers are \nolinkurl{doubao-seedream-5-0-260128} for images, \nolinkurl{doubao-seedance-2-0-260128} for videos, and \nolinkurl{doubao-seed-2-0-pro-260215} for generation/verification reasoning. Reported media settings are 720p, 16:9, with a shared stylized 3D appearance. The implementation's image-generation and video-submission limits are ten and five per shot, respectively; these limits are caps, not observed averages or evidence of equal budgets.

\paragraph{Current interface and experimental protocols.}
\Cref{sec:verification} describes the current one-node-per-turn L2 interface. The reported generation cohorts also include earlier multi-turn and consensus-review implementations with a video-edit backend. They share the node-linked verification-to-repair design; conversation and editing protocols are run-specific. The ten-image/five-video implementation caps are not substituted for historical run limits when reproducing those results.

\paragraph{Fixed-first-frame video-stage experiment.}
\Cref{tab:ablation} uses the ten-second control experiment rather than the default full-pipeline end-to-end ablation. Its protocol supplies the baseline-generated first frame to \texttt{abl1}, \texttt{abl2}, and \method{}, with no additional first-frame synthesis. The baseline uses its own keyframe selector and no SOIG video verification. Both flat configurations use event-list video verification and refinement, while \method{} uses graph-based video verification and editing. The two flat run identifiers do not denote different image-verification policies in this experiment.

For context, in the \emph{default E2E configuration}, \texttt{abl1} instead reuses a graph-verified \method{} frame, and \texttt{abl2} regenerates the frame using a flat event list. Those image-stage interventions are disabled in the control experiment reported here. Its results therefore examine video-stage verification and repair with a shared initial frame, rather than the effect of graph verification at the image stage.

\paragraph{Type C source and evaluation coverage.}
The final \method{} aggregate uses a separate Type C implementation for the 11 multi-scene cases, contributing 128 evaluated clips. Across all 35 themes, the valid report counts are 410, 419, 429, and 425 for baseline, \texttt{abl1}, \texttt{abl2}, and \method{}, respectively. Counts reflect generated outputs and report availability, including a small number of JSON failures. All conditions cover the same themes, but retained shot records are not a one-to-one matched set. The Type C source is therefore identified separately for reproducibility.

The control evaluator uses four-case, eight-shot concurrency and excludes camera-motion requirements from the physics contract. SF retains the full reference script; artifact assessment is prompt-blind and receives neither the generation prompt nor the reference script. These settings and the ten-second duration are specific to \cref{tab:ablation}, not the system comparison or the separate verification-only tool, which re-scores existing media without running generation or repair.

\section{Evaluation Protocols and Aggregation}
\label{app:evaluation}

\subsection{Benchmark Cases and Baseline Configurations}
ViMax-Bench has 12 Type A, 12 Type B, and 11 Type C themes. The system comparison uses the reported case-level aggregates; the video-stage experiment spans all 35 themes with its own retained output set. The main-result summary does not enumerate a per-system case denominator, so the benchmark size is not substituted for that denominator. The system comparison specifies a 15-second ViMax baseline, whereas the video-stage experiment uses ten-second clips. The latter is a separate generation experiment, not a duration-normalized reaggregation of the main table.

Copyright restrictions limit our evaluation of ST-Bench~\citep{zhang2025storymem} to ten cases: \texttt{artist}, \texttt{black}, \texttt{cat}, \texttt{dog}, \texttt{japan}, \texttt{phd}, \texttt{daiyu}, \texttt{jane}, \texttt{qin}, and \texttt{robinson}. We refer to this subset as ST-Bench-10. ViMax, MovieAgent, AniMaker, and \method{} are run on these same ten cases with five-second clips. The MLLM evaluator scores their generated outputs, with results reported in \cref{tab:automatic}.

ViMax, MovieAgent, and AniMaker are instantiated with a common nominal image/video backend. {AniMaker retains its MCTS-based search framework and current-shot candidate selection, with three current-shot candidates and no next-shot lookahead ($w_1=3,w_2=0$). \Cref{app:baseline-configurations} explains this setting.} The main comparison evaluates the resulting pipelines; sharing a rendering backend does not impose equal generation budgets.

\subsection{Post-Generation Evaluator}
The evaluator is \texttt{doubao-seed-2-0-pro-260215} using the \texttt{prompts\_aigc\_gaffe} rubric. The script-fidelity route uses the reference script to compile atomic requirements, record observations, and verify them. Artifact assessment is prompt-blind: neither the coarse scan nor local detail verification receives the generation prompt or reference script. The physics route separates script-derived prerequisites and rules from global physical-commonsense judgments that receive only visual evidence and timestamps. Overview sampling merges 16 uniform frames with up to eight motion-peak frames selected from a 48-frame scan, after timestamp deduplication and chronological sorting. Detail verification samples local intervals at eight fps; the maximum image dimension is 1024 pixels.

Artifact severity is 0 for a non-artifact, 1 for a defect apparent only under close/framewise inspection, 2 for a defect noticeable during ordinary viewing, and 3 for an obvious generation failure. Confirmed events receive surface, structural, or temporal categories. Event counts depend on proposal and sampling policies; they do not exhaust every defect in a clip. The evaluator is invoked separately from the control judge, so the reported score is not copied from the generation loop. However, both use the same model family and may therefore share perceptual errors when interpreting the video.

\begin{figure}[!htbp]
  \centering
  \includegraphics[width=\linewidth]{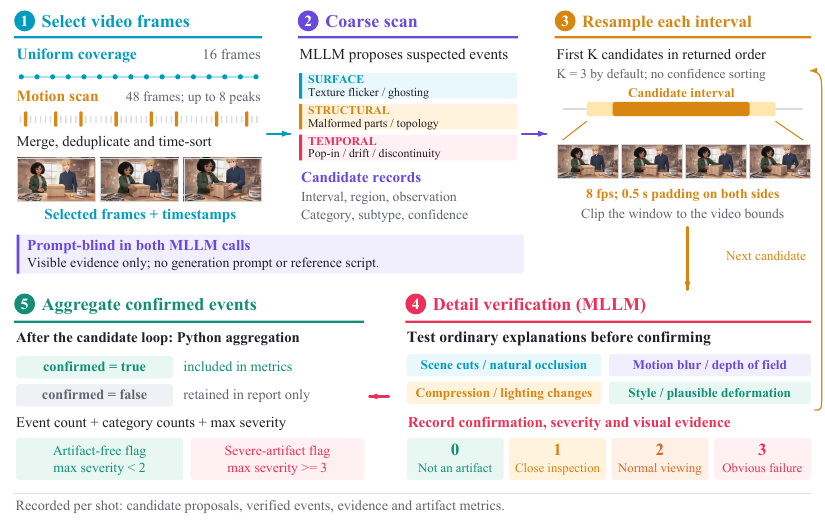}
  \caption{\textbf{Prompt-blind artifact assessment in ShotEval-MLLM.}
  Uniform and motion-peak frames are merged in temporal order for an MLLM coarse scan.
  The first $K$ proposed events are inspected using denser local frames with temporal padding.
  Neither MLLM call receives the generation prompt or reference script: the coarse scan uses video frames and timestamps, and detail verification additionally receives the suspected event.
  Detail verification tests ordinary explanations and returns confirmation, severity, and visual evidence.
  Python aggregates confirmed events into event counts, category counts, and severity-based shot flags;
  unconfirmed events remain in the report but do not contribute to these metrics.
  The displayed settings use 16 overview frames and up to eight motion peaks from a 48-frame scan, with the default detail budget $K=3$, eight-fps local sampling, and 0.5-second padding on each side.
  Real frame strips illustrate sampling rather than a recorded evaluator verdict; only a subset of sampled frames is shown.}
  \label{fig:artifact-evaluation}
\end{figure}

\Cref{fig:artifact-evaluation} summarizes the prompt-blind artifact route. Complete evaluation prompt templates appear in \cref{app:mllm-prompts}.

\FloatBarrier

\subsection{Metric Definitions}
The names below are those of the evaluation reports, not newly proposed metrics.
\begin{itemize}
 \item \textbf{SF / Atomic Fidelity}: for a shot, PASS claims divided by PASS + FAIL + UNCERTAIN claims. Uncertain claims stay in the denominator and do not count as passes.
 \item \textbf{Phys / Physics}: a 1--5 global physical-commonsense score. ViMax-Bench's system comparison averages available scores on physics-evaluable shots. ST-Bench-10 and the video-stage experiment pool available shot scores. For the latter, the physics contract excludes camera-motion requirements while retaining the other physical requirements of the shot.
 \item \textbf{Joint, ViMax-Bench}: shots with global commonsense at least 4 divided by physics-evaluable shots; it does not additionally threshold SF. An evaluable shot without a commonsense score stays in the denominator but cannot contribute to the numerator.
 \item \textbf{Joint, ST-Bench-10}: the evaluator's \texttt{joint\_semantic\_physics\_success} flags divided by all evaluated shots. The code sets this flag when the shot is physics-evaluable and its global commonsense is at least 4; no separate SF threshold is required. Nonevaluable or unscored shots contribute no success. Both remain in the all-shot denominator.
 \item \textbf{Joint, video-stage experiment}: the proportion of physics-evaluable shots with global commonsense at least 4, matching the ViMax-Bench main-table definition. This differs from some exports' all-shot success flag, whose denominator includes nonevaluable shots.
 \item \textbf{Severe\%}: the percentage of evaluated shots with at least one confirmed severity-3 event. ViMax-Bench reports the mean of case-level percentages; ST-Bench-10 pools evaluated shots. In \cref{tab:ablation}, case-level percentages are macro-averaged over the 35 themes.
 \item \textbf{Video Gen / Shot and Video Edit / Shot}: counts of video generation and editing operations divided by selected output shots, called \emph{promoted shots} in the run logs. This denominator counts selected outputs, rather than only shots with valid post-generation evaluation reports.
 \item \textbf{Video Try / Shot}: the sum of generation and editing operations per selected shot in \cref{tab:ablation}. The single-generation baseline is 1.00; editing also counts toward the total.
\end{itemize}

ViMax-Bench's main comparison uses case-level aggregation; ST-Bench-10 pools evaluated shots. In \cref{tab:ablation}, SF and Physics pool valid shot scores, Joint conditions on physics evaluability, and Severe\% averages theme-level percentages. Values are compared among configurations within the same experimental setting, since these aggregation rules weight cases and shots differently.

\subsection{Human Assessment}
\label{app:human}
\begin{figure}[p]
  \centering
  \includegraphics[width=\linewidth,height=0.82\textheight,keepaspectratio]{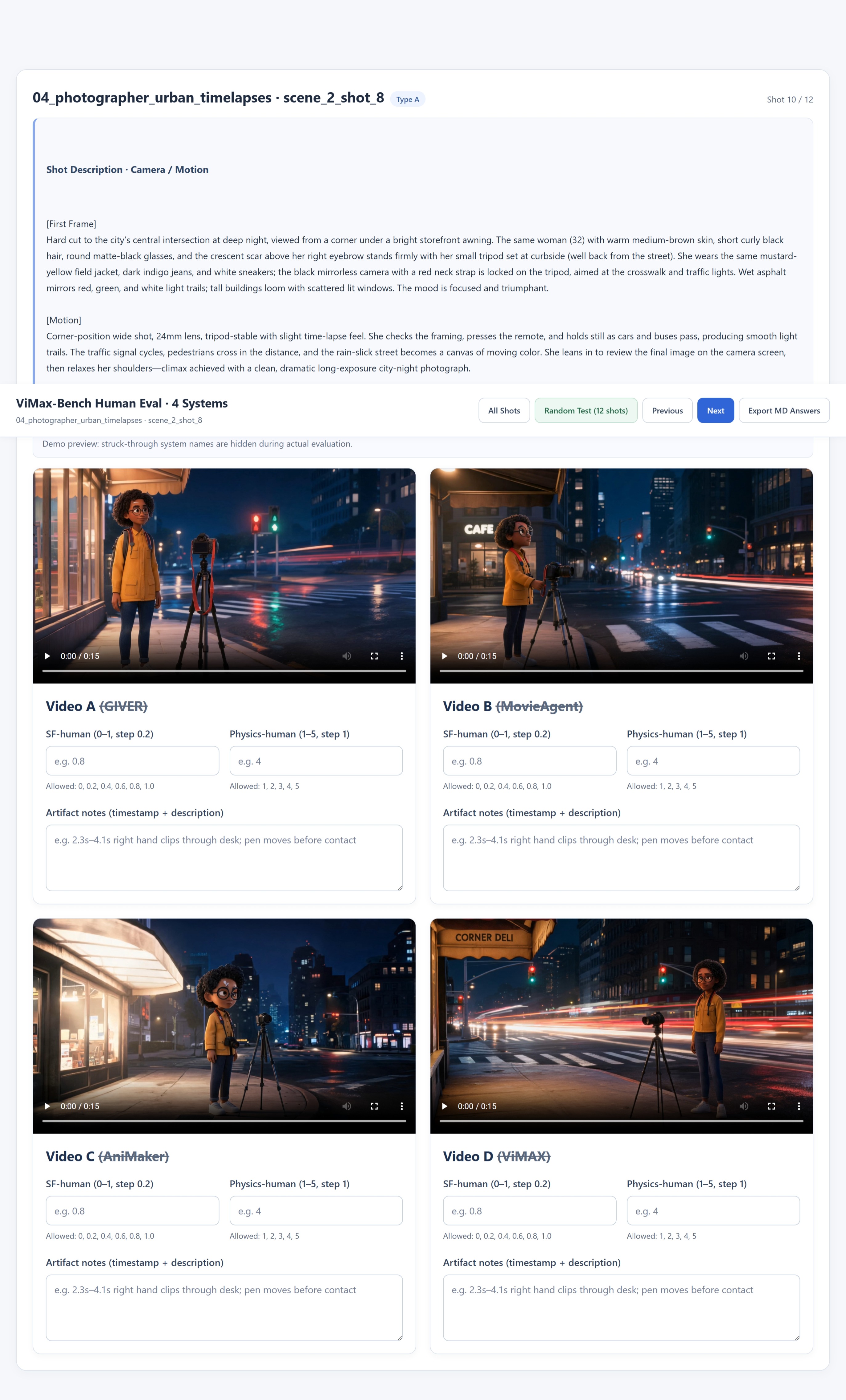}
  \caption{\textbf{Human-evaluation interface, shown with illustrative system labels.}
  Participants viewed the reference shot description and four anonymized videos, rated script fidelity (SF-human, 0--1 in increments of 0.2) and physical plausibility (Physics-human, 1--5 in increments of 1), and could record artifact timestamps and descriptions.
  The struck-through system names are annotations added solely for this illustration; these names were entirely hidden during the actual evaluation.}
  \label{fig:human-evaluation-interface}
\end{figure}

Each displayed shot group contains four anonymized system videos and the corresponding requested content. Raters assign script fidelity and physical-plausibility scores on their original scales. Video order within shots and shot-group order are randomized; codenames conceal system identities. \Cref{fig:human-evaluation-interface} shows the ViMax-Bench rating interface, with system names added only for illustration.

ViMax-Bench has 16 raters, 12 groups per rater, and four system videos per group: 768 valid system--shot ratings. The combined sample contains 148 unique videos over 37 shots. For ST-Bench-10, 25 raters assess 24 shot groups, each containing one video per system ($24\times4=96$ unique videos across four systems), producing 2,400 submitted ratings. Seventy-eight records contain at least one score outside SF's [0,1] or physics' [1,5] range and are excluded. The remaining counts are 579 for ViMax, 578 for MovieAgent, 582 for AniMaker, and 583 for \method{}. \Cref{tab:human} preserves the original system means, without rescaling physics to [0,1].

The reporting script computes the system means from valid rating records; equivalently, these are per-video human means weighted by the number of valid ratings. For correlation analysis, repeated ratings are first averaged within each system--shot video and then paired with that video's automated score. Original-scale SF/physics Pearson correlations are .022/.483 for ViMax-Bench and .487/$-.026$ for ST-Bench-10. Thus aggregate system rankings and video-level agreement describe different aspects of the evaluation. We report the observed system means without a statistical-significance claim, since repeated ratings of the same video are not independent samples.

\FloatBarrier
\subsection{Accounting Boundaries}
\label{app:accounting}
System-comparison \method{} video-request means are 1.51 on ViMax-Bench and 1.88 on ST-Bench-10. ViMax and MovieAgent records generally stop after the first successful video request; configured transport retry limits do not mean three videos were generated.

The video-stage report separates Video Gen / Shot from Video Edit / Shot, then adds them as Video Try / Shot. Reported component sums are 1.000, 2.856, 2.910, and 2.183 for baseline, \texttt{abl1}, \texttt{abl2}, and \method{}. The last value is about 23.6\% and 25.0\% below the two flat configurations, respectively. Rounded totals in \cref{tab:ablation} are 1.00, 2.86, 2.91, and 2.18. The baseline edit entry follows the summary's dash notation; its total describes a single image-to-video operation.

Video generation and editing may have different prices, so Video Try measures media operations rather than a verification-inclusive monetary budget. Asset preparation, requirement compilation, judge calls, prompt refinement, failures, and caching contribute additional costs. {The additional analysis in \cref{app:verification-cost} reports graph-compilation and verification expenditure separately. Wall-clock latency and human-inspection time are not included in those dollar figures.}

\subsection{Paired Analysis of Retried Shots}
\label{app:retry-analysis}

\paragraph{Candidates and evaluation.}
This analysis compares the first video candidate with the final selected output for shots with more than one recorded video attempt. The first candidate is \texttt{video\_000/candidate.mp4}; the final output is the shot's selected \texttt{video.mp4}, not necessarily its last generated candidate. The original benchmark script is the reference for both outputs. The evaluation uses the AIGC artifact rubric and excludes camera-motion requirements from the physics contract. Only pairs with numeric SF and Physics scores for both outputs enter the aggregate. This yields 177 shots from 34 ViMax-Bench cases. The fifteen-second setting is separate from the ten-second experiment in \cref{tab:ablation}; scores from these two experiments are not pooled.

\paragraph{Historical run selection.}
The source export selects historical runs using case-level Physics scores. If the selected source is a batch run, the analysis uses that run's candidate history. If it is a packaged output bundle, the resolver selects the highest-Physics batch with multi-attempt shots; its fallback selects the batch containing the most multi-attempt shots. Selection therefore uses post-generation evaluation, rather than a score-independent assignment of runs. We report the resulting paired changes descriptively: they do not identify the average causal effect of retrying over all runs, or separate feedback-guided repair from an equal-budget random-resampling alternative. This selection is distinct from the controller's within-run choice of its final candidate during video generation.

\paragraph{Paired scores and baseline intersection.}
Means give equal weight to each retained shot, not to each case. Across all 177 pairs, SF changes from .7496 to .7619 (a mean difference of .0123), while Physics changes from 3.3729 to 3.6780 (a mean difference of .3051). Physics improves on 70 shots, is unchanged on 67, and decreases on 40. The corresponding SF counts are 103, 2, and 72. The median Physics change is zero, so a positive average does not imply that most individual shots improve. Intermediate candidates are not evaluated in this comparison; it does not establish a monotonic improvement with attempt count or determine how many attempts a shot needs.

The baseline comparison matches the case and shot identifiers and retains only shots with all five outputs' SF and Physics scores. Its intersection contains 146 shots from 33 cases; all methods in \cref{fig:retry-analysis} use this same subset. SF means are .7578, .7572, and .7701 for ViMax, MovieAgent, and AniMaker, versus .7488 and .7651 for \method{} Initial and Final. Physics means are 3.5685, 3.4932, and 3.5205, versus 3.3288 and 3.6164. These baseline values contextualize the selected subset, rather than replace the full-benchmark system comparison. We make no statistical-significance or verification-inclusive cost claim from these paired scores on the matched shot subset.

\section{Illustrative Media and Figure Provenance}
\label{app:figures}

\Cref{fig:teaser} uses real images and frames from a historical cooking-scene repair trace. The initial pair displays a count mismatch and its corrected realization. Video candidates are shown at matched timestamps, exposing unexplained food appearance in the earlier clip. Rectangles are annotations of the observed defects, not learned localization outputs or pixel-level masks used for repair.

\Cref{fig:pipeline} uses a historical ViMax-Bench gardening trace. The expanded contract has category-grouping branches and thirteen logical prerequisite links; the former visualize organization, whereas the latter define dependencies. The recorded A1 action remains satisfied at 6.0 seconds. At 9.0 seconds, AIGC-CH1 identifies a missing garden trowel, prompting revision of the motion field while retaining the first frame. The selected subsequent clip records a pass and matches the saved evaluation video. The matched close-ups use the same crop window and timestamp in the source frames.

Both traces predate the one-node-per-turn protocol. The gardening example used multi-turn requirement review and two full-video continuity reviews. Frames are unretouched except for documented crops and overlays; \cref{tab:ablation} reports the separate quantitative video-stage experiment.

\Cref{fig:qualitative-comparison} uses two ViMax-Bench tasks from the blinded comparison bundle: a mentor--student woodworking interaction and a wine-counter interaction. All four systems are shown at the same timestamps within each task. The woodworking frames are at 0, 6, 10, and 12 seconds; the wine-counter frames are at 0, 1.5, 11, and 14 seconds. The 32 main frames retain their full field of view. Five detail insets enlarge marked regions from the same frames; they do not introduce additional observations from different timestamps. Red marks observed defects, gray marks an earlier comparison state, and green marks the illustrated \method{} behavior. Frames, insets, and timestamps are not altered to simulate a repair. The cases were selected to illustrate geometric and temporal failure modes, not sampled to estimate their frequency. Unmarked frames do not establish that a clip is artifact-free.

\section{Additional Context for Verification and Evaluation}
\label{app:context}

\paragraph{Where additional computation is spent.}
The three mechanisms in \cref{sec:related} distinguish control decisions rather than agent names. Related single-clip optimization methods further clarify the scope of our efficiency experiment. Video-T1 searches generation trajectories and prunes branches~\citep{liu2025videot1}, whereas SketchVerify evaluates inexpensive rendered motion sketches before full synthesis~\citep{huang2025sketchverify}. CCC supplies structured multimodal critique~\citep{gu2025ccc}. VPO instead learns a prompt optimizer using text- and video-level feedback~\citep{cheng2025vpo}; its training procedure is distinct from a test-time local repair loop. These approaches allocate computation to different objects: trajectories, plans, critiques, or learned prompts. Our reported operation count measures the generation and editing performed by the video-stage controller with a fixed first frame. It does not compare internal compute or establish an optimum over all search strategies.

\paragraph{From specific feedback to a repair target.}
Decomposed feedback has precedents in image generation. Divide, Evaluate, and Refine uses assertion-level VQA feedback to improve alignment~\citep{singh2023divide}; Rich Human Feedback identifies problematic regions and misrepresented prompt terms~\citep{liang2024richhf}. VisionDirector checks structured visual goals and incorporates staged edits and rollback~\citep{chu2026visiondirector}. These methods motivate separating a diagnosis from a single aggregate score. In \method{}, the additional requirement is to keep the shot-level check identifiable across media candidates and connect it to an allowed text-field update. The graph is a record of requirements and prerequisites, not a guarantee that the visual judge is correct. The marked regions explain observed defects; they are not spatial masks passed to the generator for editing.

\paragraph{Correction requires evidence.}
Self-Refine demonstrates the usefulness of iterative language feedback~\citep{madaan2023selfrefine}, while studies of self-correction show that improvements depend on the feedback available and the task~\citep{huang2024cannotselfcorrect,kamoi2024selfcorrection}. We do not assume that repeating a judgment will make it correct. A new candidate must provide new visual evidence for the same requirement, and uncertain evidence stays uncertain unless a later check resolves it. Similarly, work on reward-model overoptimization shows why maximizing a proxy alone is not sufficient evidence of improved task performance~\citep{gao2023rewardoveroptimization}. The external quality tables and blinded ratings are therefore separate from the diagnostic score used to accept a candidate.

\paragraph{What a benchmark measures.}
Broad video-quality suites such as VBench and EvalCrafter cover multiple perceptual and alignment dimensions~\citep{huang2024vbench,liu2024evalcrafter}; T2V-CompBench emphasizes compositional requirements~\citep{sun2025t2vcompbench}. Story-oriented benchmarks target another level of assessment: ViStoryBench evaluates coherent story visualization, MSVBench studies multi-shot video evaluation, and EntityBench focuses on entity consistency across longer shot sequences~\citep{zhuang2026vistorybench,shi2026msvbench,he2026entitybench}. SLVMEval uses controlled synthetic perturbations to examine long-video evaluation metrics~\citep{matsuda2026slvmeval}. Our ViMax-Bench and ST-Bench-10 experiments use multi-shot tasks, but the reported artifact and physics metrics evaluate individual clips. Improvements in these clip-level metrics do not, by themselves, establish better pacing or cross-shot continuity across the complete assembled story.

\paragraph{System rankings and individual judgments.}
Automatic evaluation is useful for coverage, but its behavior depends on the judge and the questions it is asked. Studies of text--image metrics and video-language judges examine these reliability issues directly~\citep{ross2024goodmetric,liu2025reliablejudge}. MINERVA identifies temporal localization and visual perception as important sources of video-reasoning errors, while Video-MME evaluates understanding across video lengths and modalities~\citep{nagrani2025minerva,fu2025videomme}. This motivates reporting both automated and human assessment and preserving their distinct units of analysis. The system means in \cref{tab:human} summarize viewer ratings; the correlations in \cref{app:human} ask whether two evaluators agree on individual system--shot videos. Agreement on the best system does not make these statistics equivalent.

%% file: appendix_mllm_prompts.tex
\clearpage
\begingroup
\raggedbottom
\section{Prompts for MLLM-Based Evaluation}
\label{app:mllm-prompts}

This section provides the prompt templates used by ShotEval-MLLM to assess script fidelity, visible artifacts, and physical commonsense after generation. They are separate from GIVER's repair prompts. The evaluator uses \texttt{doubao-seed-2-0-pro-260215}. The system and user message bodies below preserve the original instructions, scoring rubrics, and output schemas. Braced placeholders denote runtime inputs; role headings and line wrapping are presentation only.

\paragraph{Inputs and prompt visibility.}
Overview images combine 16 uniformly sampled frames with up to eight motion peaks selected from a 48-frame scan, after timestamp deduplication and chronological sorting. Images are attached separately from the text. The placeholder \texttt{\{frame\_manifest\}} lists their image indices and timestamps. Neither artifact call nor the global physical-commonsense call receives the generation prompt or reference script. Script-fidelity observation and physics-contract construction do use the reference requirements. Thus \emph{prompt-blind} refers to withholding the generation prompt or reference script, not to withholding the evaluation instructions.

\begin{center}
\small
\begin{tabular}{@{}p{.12\linewidth}p{.49\linewidth}p{.31\linewidth}@{}}
\hline
Template & Function & Visual input \\
\hline
SF-1 & Compile the reference script into atomic claims & None \\
SF-2 & Observe evidence for one claim & Overview frames \\
SF-3 & Judge the claim from observations and dependencies & None; text evidence \\
AR-1 & Propose visible artifact intervals & Overview frames \\
AR-2 & Confirm an event and assign severity & Dense local frames \\
PH-1 & Compile applicability and physical prerequisites & None \\
PH-2 & Determine whether prerequisites visibly occur & Overview frames \\
PH-3 & Score global physical commonsense & Overview frames \\
\hline
\end{tabular}
\end{center}

The templates cover the model outputs used by the reported metrics. Joint and Severe\% are computed from those outputs, not elicited through additional rating prompts. Separate key-phenomenon, event-order, rule-grounding, and pairwise interfaces are not reproduced here because they do not enter these reported metric definitions.

\subsection{Script Fidelity}
\label{app:prompts-sf}
SF-1 constructs the atomic claims from the reference script. SF-2 receives that script, one claim, and sampled images; SF-3 receives the observation record and prerequisite results, without an additional image attachment. Claims are processed after their prerequisite results are available. The shot-level SF score is the number of PASS claims divided by PASS, FAIL, and UNCERTAIN claims together. An unresolved claim is not counted as a pass.

\Needspace{8\baselineskip}
\textbf{SF-1 input.} \texttt{\{script\}} is the reference shot script. The call is text-only; its output supplies the claims used by SF-2 and SF-3.

\begin{giverpromptbox}{SF-1: Script contract construction}{giverPromptBlue}
[System message]
You convert a generated-video shot script into atomic, visually verifiable claims.
Return JSON only. Do not judge a video and do not add requirements that are not stated or visually implied.
Each claim must contain exactly one fact. Psychological states must be translated only when the script gives a visible manifestation.

[User message template]
Compile the following shot script into a semantic contract.

SHOT SCRIPT:
{script}

Allowed claim types:
entity, attribute, number, action, interaction, spatial, temporal_order, state_change, end_state, camera.

Rules:
1. Every claim must be independently visually decidable.
2. Mark core=true only for facts essential to the shot's main content.
3. Add dependencies using claim IDs when a claim cannot be evaluated unless another claim passes.
4. Do not include physical plausibility judgments here. Only encode what should visibly occur.
5. Do not infer exact colors, counts, directions, camera motion, or object states unless specified.
6. Use concise declarative English claims.

Return this schema:
{
  "shot_summary": "...",
  "claims": [
    {
      "id": "E1",
      "type": "entity",
      "claim": "A woman is visible.",
      "core": true,
      "dependencies": []
    }
  ]
}
\end{giverpromptbox}

\Needspace{8\baselineskip}
\textbf{SF-2 inputs.} \texttt{\{script\}}, \texttt{\{claim\}}, and \texttt{\{claim\_id\}} specify the reference and target claim. \texttt{\{frame\_manifest\}} corresponds to the attached overview images. This call describes evidence rather than returning the final verdict.

\begin{giverpromptbox}{SF-2: Evidence observation}{giverPromptBlue}
[System message]
You are the observation stage of an evidence-grounded video-script evaluator.
You must describe only visible evidence relevant to one atomic claim. Do not issue the final PASS/FAIL verdict.
The images are ordered video frames. Their timestamps are provided in the prompt.
Never infer hidden events between sampled frames. Use UNCLEAR when evidence is insufficient.
Return JSON only.

[User message template]
SHOT SCRIPT:
{script}

ATOMIC CLAIM:
{claim}

FRAME TIMESTAMPS:
{frame_manifest}

Locate evidence relevant to the claim. Describe only what can be visibly verified.
For actions and temporal claims, identify an evidence interval when possible.

Return:
{
  "claim_id": "{claim_id}",
  "observability": "CLEAR|PARTIAL|UNCLEAR",
  "observations": [
    {
      "start_time": 0.0,
      "end_time": 0.0,
      "description": "visible fact only"
    }
  ],
  "missing_evidence": "short explanation or null"
}
\end{giverpromptbox}

\Needspace{8\baselineskip}
\textbf{SF-3 inputs.} The claim record supplies \texttt{\{claim\}}, \texttt{\{claim\_type\}}, \texttt{\{core\}}, and \texttt{\{claim\_id\}}. Prerequisite verdicts and evidence populate \texttt{\{dependency\_results\}}; the SF-2 output supplies \texttt{\{observation\}}. This call receives text only, with no image attachment.

\begin{giverpromptbox}{SF-3: Claim verification}{giverPromptBlue}
[System message]
You are the verification stage of an evidence-grounded video-script evaluator.
Decide whether one atomic claim is satisfied using only the supplied observations and claim dependencies.
No evidence means no PASS. Return JSON only.

[User message template]
ATOMIC CLAIM:
{claim}

CLAIM TYPE:
{claim_type}

CORE CLAIM:
{core}

DEPENDENCY RESULTS:
{dependency_results}

OBSERVATION STAGE OUTPUT:
{observation}

Decision policy:
- PASS: clear visible evidence directly satisfies the entire atomic claim.
- FAIL: clear evidence contradicts the claim, or a required event/entity is absent despite adequate visibility.
- UNCERTAIN: sampling, occlusion, blur, ambiguity, or failed dependencies prevent a reliable judgment.
- A temporal-order claim can PASS only when all required events are themselves observed and their order is clear.
- An attribute/action claim cannot PASS when its referenced entity dependency failed.

Return:
{
  "claim_id": "{claim_id}",
  "verdict": "PASS|FAIL|UNCERTAIN",
  "confidence": 0.0,
  "evidence": [{"start_time": 0.0, "end_time": 0.0, "description": "..."}],
  "failure_reason": "wrong_entity|missing_entity|wrong_attribute|missing_action|wrong_action|missing_interaction|wrong_spatial_relation|wrong_event_order|wrong_state|camera_mismatch|insufficient_evidence|null"
}
\end{giverpromptbox}

\Needspace{14\baselineskip}
\subsection{Prompt-Blind Artifact Assessment}
\label{app:prompts-artifact}
AR-1 proposes candidate intervals from the overview frames. AR-2 receives one candidate and dense images from its local interval, with timestamps but no reference script. The detail defaults inspect the first three returned candidates at eight fps, with 0.5-second padding on each side and the window clipped to the video bounds. Candidates are processed in returned order, not re-ranked by confidence. Only confirmed events enter artifact aggregation. A shot is severe when its maximum confirmed severity is at least 3; Severe\% aggregates these shot flags according to the experiment's stated weighting. Both stages judge the visible clip without access to the reference script.

\Needspace{8\baselineskip}
\textbf{AR-1 input.} The attached overview images and \texttt{\{frame\_manifest\}} are the only shot-specific inputs. No script-derived requirements are supplied.

\begin{giverpromptbox}{AR-1: Artifact coarse scan}{giverPromptRed}
[System message]
You detect visible generative artifacts in a video without access to its prompt or script.
Judge only intrinsic visual/temporal defects. Do not penalize missing requested content, unusual but possible events, intentional style, normal motion blur, depth of field, compression, or camera cuts.
The images are ordered frames with timestamps. Return JSON only.

[User message template]
FRAME TIMESTAMPS:
{frame_manifest}

Scan for candidate artifact events in these categories:
- SURFACE: texture boiling, local flicker, ghosting, unexplained color/exposure jumps, smeared or melting textures, repeated texture, contaminated boundaries.
- STRUCTURAL: malformed face/hand/body/object, extra or missing parts, merging/splitting, unstable topology, incorrect occlusion boundary, foreground-background adhesion, reflection/shadow geometry mismatch.
- TEMPORAL: abrupt appearance/disappearance without cut or occlusion, identity/shape drift, teleportation, discontinuous local motion, incompatible foreground/background motion, non-rigid background warping during camera motion.

Identify only candidate intervals worth detailed verification. Avoid concluding from a single low-quality frame.

Return:
{
  "candidates": [
    {
      "category": "SURFACE|STRUCTURAL|TEMPORAL",
      "subtype": "short_snake_case_label",
      "start_time": 0.0,
      "end_time": 0.0,
      "region": "visible region or object",
      "observation": "what appears suspicious",
      "confidence": 0.0
    }
  ]
}
\end{giverpromptbox}

\Needspace{8\baselineskip}
\textbf{AR-2 inputs.} One AR-1 event record supplies \texttt{\{candidate\}}. Dense local images are attached with their timestamps in \texttt{\{frame\_manifest\}}. The output reports confirmation, severity, and the alternative explanations tested for that event.

\begin{giverpromptbox}{AR-2: Artifact detail verification}{giverPromptRed}
[System message]
You verify one suspected generative artifact using densely sampled frames from a short interval.
You must actively test ordinary explanations before confirming an artifact. Return JSON only.

[User message template]
SUSPECTED EVENT:
{candidate}

DENSE FRAME TIMESTAMPS:
{frame_manifest}

Determine whether the suspected event is a genuine visible artifact.
Explicitly test these alternatives when relevant:
scene cut, natural occlusion, motion blur, depth of field, compression, lighting change, intentional stylization, physically plausible deformation.

Severity rubric:
0 = not an artifact.
1 = only visible under close frame-by-frame inspection.
2 = noticeable during normal viewing but does not dominate the shot.
3 = immediately obvious and strongly exposes the video as generated or broken.

Return:
{
  "confirmed": true,
  "category": "SURFACE|STRUCTURAL|TEMPORAL",
  "subtype": "...",
  "start_time": 0.0,
  "end_time": 0.0,
  "region": "...",
  "bbox_normalized": [0.0, 0.0, 1.0, 1.0],
  "severity": 0,
  "confidence": 0.0,
  "observation": "specific visible evidence across frames",
  "alternative_explanations": [
    {"name": "motion blur", "accepted": false, "reason": "..."}
  ]
}
\end{giverpromptbox}

\Needspace{17\baselineskip}
\subsection{Physical Commonsense and Evaluability}
\label{app:prompts-physics}
PH-1 derives a physics contract from the reference script, including an applicability flag and semantic prerequisites. In experiments configured to exclude camera-motion requirements, the script is filtered before this call; the template itself is unchanged. PH-2 checks whether those prerequisites visibly occur. It does not score physical correctness. For an applicable contract, PH-3 independently scores visible physical commonsense from frames and timestamps only, even if PH-2 cannot establish evaluability. An inapplicable contract yields no physical-commonsense score.

Phys uses the numeric PH-3 score with the selection and aggregation rules specified for each experiment. Joint requires \texttt{physics\_evaluable=true} and a numeric PH-3 score of at least 4; it has no additional SF threshold. Its denominator remains experiment-specific: physics-evaluable shots for the ViMax-Bench and fixed-first-frame reports, and all evaluated shots for ST-Bench-10. The auxiliary phenomenon, event-order, and rule fields in PH-1's original schema are preserved verbatim; they do not define an extra condition for these reported Joint scores.

\Needspace{8\baselineskip}
\textbf{PH-1 input.} \texttt{\{script\}} is the reference script, or its camera-filtered form for the specified experiments. No generated frames are supplied to this text-only compilation call.

\begin{giverpromptbox}{PH-1: Physics contract construction}{giverPromptGold}
[System message]
You convert a shot script into an observable physics contract for later video evaluation.
Return JSON only. Focus on ordinary real-world physical behavior, not mathematical derivations.
AIGC CONTEXT: The video that will be judged is AI-generated and may contain gaffes (artifacts). Include ordinary body-integrity / impenetrability / object-persistence rules so later judges can catch extra limbs, fusion, floating, and sudden replacements.

[User message template]
Build a physics contract for the following shot script.
The evaluated video is AIGC and may have artifacts; keep body-integrity and impenetrability rules even for simple actions.

SHOT SCRIPT:
{script}

Rules:
1. Set applicable=false when no meaningful motion, interaction, material response, fluid behavior, optics, or gravity-related event is expected.
2. Semantic preconditions state what must visibly occur before physics can be evaluated.
3. Expected key phenomena must be directly observable in video.
4. Expected event order must describe causal temporal order.
5. Applicable rules should be short neutral statements. Prefer categories: gravity, inertia, momentum, collision, impenetrability, material_response, fluid_dynamics, buoyancy, reflection, shadow, object_persistence, kinematic_continuity.
6. Forbidden outcomes must be concrete visible failures.
7. Do not over-specify outcomes when multiple real-world outcomes would all be plausible.

Return this schema:
{
  "applicable": true,
  "semantic_preconditions": ["..."],
  "expected_key_phenomena": ["..."],
  "expected_event_order": ["event 1", "event 2"],
  "applicable_rules": [
    {"category": "impenetrability", "rule": "Objects should not pass through each other during contact."}
  ],
  "forbidden_outcomes": ["..."]
}
\end{giverpromptbox}

\Needspace{8\baselineskip}
\textbf{PH-2 inputs.} \texttt{\{preconditions\}} comes from PH-1. \texttt{\{frame\_manifest\}} lists the attached overview images. This stage determines \texttt{physics\_evaluable}, not the 1--5 quality score.

\begin{giverpromptbox}{PH-2: Semantic prerequisites}{giverPromptGold}
[System message]
You evaluate whether the visible prerequisites for a physical event occurred in a video.
This stage does not judge whether the physics is correct. Return JSON only.
AIGC CONTEXT: The clip is AI-generated video and may contain visible gaffes (artifacts). Judge only pixels; do not invent off-screen body parts to satisfy a precondition.

[User message template]
SEMANTIC PRECONDITIONS:
{preconditions}

FRAME TIMESTAMPS:
{frame_manifest}

For each precondition, judge whether it is visibly satisfied.
Return:
{
  "items": [
    {
      "precondition": "...",
      "verdict": "PASS|FAIL|UNCERTAIN",
      "evidence_time": [0.0, 0.0],
      "observation": "..."
    }
  ],
  "physics_evaluable": true,
  "reason": "..."
}
\end{giverpromptbox}

\Needspace{8\baselineskip}
\textbf{PH-3 input.} Only overview images and \texttt{\{frame\_manifest\}} are supplied. Neither the script, the physics contract, nor the PH-2 verdict is included in this prompt.

\begin{giverpromptbox}{PH-3: Global physical commonsense}{giverPromptGold}
[System message]
You are a prompt-blind evaluator of visible physical commonsense in a video.
Judge only the behavior shown in the frames. Do not assume intended events from any script.
Return JSON only.
AIGC CONTEXT: The video is AI-generated and may have production gaffes (artifacts). Extra limbs, interpenetration, floating objects, melting geometry, and unexplained appearance/disappearance are physical violations, not style. Do not give the benefit of the doubt for off-screen explanations.

[User message template]
FRAME TIMESTAMPS:
{frame_manifest}

This clip is AIGC and may contain artifacts. Considering ordinary real-world behavior, rate the physical plausibility and naturalness of the visible process. Penalize visible AIGC gaffes.

Rubric:
1 = severe and obvious violations of everyday physics.
2 = multiple clear implausible behaviors.
3 = understandable process with one or more significant anomalies.
4 = largely plausible with only minor anomalies.
5 = natural and physically credible throughout the observable process.

Separately consider motion continuity, object persistence, contact/interaction response, gravity/material behavior when applicable, and causal timing.

Return:
{
  "score": 1,
  "confidence": 0.0,
  "observed_process": "neutral description",
  "violations": [
    {"type": "...", "time": [0.0, 0.0], "description": "..."}
  ]
}
\end{giverpromptbox}

\par\endgroup

%% file: appendix_rebuttal.tex
\clearpage
\begingroup

\hypersetup{linkcolor=refblue,citecolor=refblue,urlcolor=refblue}
\captionsetup{font={small,color=black},labelfont={bf,color=black},textfont={color=black}}

\section{Additional Experiments and Verification Evidence}
\label{app:rebuttal}

This appendix adds a three-candidate budget comparison, human assessment of verifier decisions, measured compilation and verification costs, and examples showing how diagnosis and repair work. It also examines character continuity across a shot transition. The main-result tables describe the original experiments; the additional studies are presented separately below.

\subsection{Baseline Settings and Experimental Controls}
\label{app:baseline-configurations}

\paragraph{Common task and method-specific prompts.}
All methods receive the same scene and action requirements and use the standardized rendering backend in \cref{sec:setup}. Each method turns these requirements into its own generation prompts, including its formatting and style instructions. In the end-to-end comparison, each pipeline generates its own first frame. Fidelity is scored against the original task requirements throughout generation and repair. Artifact and global physical-commonsense judgments use visual evidence and timestamps and are prompt-blind.

\paragraph{AniMaker configuration.}
ViMax and MovieAgent use standardized production-script inputs and generation backends, with their video-search modules otherwise unchanged. AniMaker uses its MCTS-based search framework and score-based selection among three current-shot candidates. We disable Next-Shot Lookahead (NSL) by setting $w_1=3$ and $w_2=0$.

This setting reflects both the task and the generation budget. ViMax-Bench includes deliberate cuts across scenes, times, and camera viewpoints. The appropriate continuity between adjacent shots therefore depends on the story. NSL also synthesizes next-shot videos and evaluates their quality and continuity. With $w_1=3,w_2=3$, a search step can generate three future candidates in addition to three current-shot candidates. Before candidate reuse, this approximately doubles generation at that step. Our configuration concentrates the budget on satisfying each prescribed shot while preserving current-shot selection.

\begin{table}[!htbp]
\centering\small
\setlength{\tabcolsep}{5pt}
\caption{\textbf{What is controlled in each experiment.} Different rows answer different experimental questions.}
\label{tab:rebuttal-controls}
\begin{tabular}{@{}>{\raggedright\arraybackslash}p{.23\linewidth}>{\raggedright\arraybackslash}p{.35\linewidth}>{\raggedright\arraybackslash}p{.36\linewidth}@{}}
\toprule
Experiment & Common conditions & Purpose \\
\midrule
Main comparisons, \cref{tab:automatic,tab:human} & Original task requirements and rendering backend; each pipeline supplies its own first frame. & Quality of complete pipelines. ST-Bench-10 uses 5-second clips; ViMax-Bench uses 15-second clips. \\
Video-stage ablation, \cref{tab:ablation} & Baseline-generated first frame; 35 themes; 10-second clips; run-specific details in \cref{app:versions}. & Video verification and repair with fixed first-frame conditioning. \\
Three-candidate comparison, \cref{tab:three-candidate} & ViMax-Bench tasks; at most three current-shot generation candidates; unchanged evaluation definitions. & Whether quality advantages persist when baselines receive additional candidates. \\
Dependency replays, \cref{app:edge-replays} & Same video, atomic checks, answers, and observations; only edges change. & The contribution of dependencies to diagnosis and repair-target assignment. \\
\bottomrule
\end{tabular}
\end{table}

The inputs for \cref{fig:qualitative-comparison} confirm that the compared methods receive common task requirements. The woodworking task specifies a rectangular sanding block and a mentor adjusting the workpiece to keep it flat. The warped workpiece is a generation error relative to that requirement. The wine task specifies bottle placement, corkscrew placement, and the passing guest for all methods. These comparisons assess how well each pipeline realizes the assigned task.
\FloatBarrier

\subsection{Comparison under a Three-Candidate Generation Budget}
\label{app:three-candidate}

\paragraph{Candidate generation and selection.}
The original evaluated ViMax and MovieAgent configurations produce one successful video per shot. To test additional sampling, we augment each with three current-shot candidates and score-based best-of-three selection. The selector is an AniEval-based adaptation of the candidate evaluator used by AniMaker~\citep{shi2025animaker}. This supplies a defined selection rule when ViMax and MovieAgent receive additional generation opportunities. AniEval evaluates candidates; the surrounding generation loop provides the repeated attempts. AniMaker uses its three-current-candidate configuration described above.

For \method{}, we apply the three-candidate cap retrospectively to its existing generation runs. Selection uses the best of the first three available candidates; shots with fewer candidates keep their selected output. We report this analysis separately from the original five-attempt configuration. The shared budget is a maximum of three video-generation candidates per shot. Monetary cost and total computation may differ between methods.

\begin{table}[!htbp]
\centering\small
\setlength{\tabcolsep}{4pt}
\caption{\textbf{ViMax-Bench quality with additional baseline sampling.} The upper block shows the original results; the lower block compares three-candidate configurations. Bold denotes the best result in the lower block. Mean and maximum count video-generation attempts per shot; \cref{tab:ablation} counts generation-plus-edit operations.}
\label{tab:three-candidate}
\begin{tabular}{@{}llrrrrrr@{}}
\toprule
System & Selection & SF $\uparrow$ & Phys $\uparrow$ & Joint $\uparrow$ & Severe\% $\downarrow$ & Mean & Max \\
\midrule
ViMax & Original & .769 & 3.76 & .512 & 3.50 & 1.00 & 1 \\
MovieAgent & Original & .753 & 3.79 & .531 & 4.20 & 1.00 & 1 \\
\method{} & Original & .780 & 4.01 & .627 & 0.70 & 1.51 & 5 \\
\midrule
ViMax & AniEval best-of-3 & .774 & 3.73 & .566 & 3.84 & 3.00 & 3 \\
MovieAgent & AniEval best-of-3 & .766 & 3.82 & .578 & 4.47 & 3.00 & 3 \\
AniMaker & Current-shot best-of-3 & .768 & 3.65 & .515 & 5.70 & 3.00 & 3 \\
\method{} & First-three cap & \textbf{.775} & \textbf{3.92} & \textbf{.595} & \textbf{2.37} & \textbf{1.42} & 3 \\
\bottomrule
\end{tabular}
\end{table}

\paragraph{Higher quality with fewer generation attempts.}
\method{} leads all four reported quality metrics in the three-candidate comparison. Its physical-plausibility score is 3.92, compared with 3.82 for the strongest baseline, and Joint is .595 versus .578. The severe-artifact rate is 2.37\%, compared with the best baseline rate of 3.84\%, a 38.3\% relative reduction. \method{} also averages 1.42 generation attempts per shot, compared with 3.00 for the baselines, using 52.7\% fewer attempts. Its advantage therefore persists when competing methods receive three generation opportunities.

Additional sampling improves SF and Joint for both ViMax and MovieAgent, while their severe-artifact rates increase slightly. The original five-attempt \method{} setting achieves still lower Severe\% (0.70) and higher mean quality than the three-candidate cap, averaging 1.51 attempts compared with 1.42. These results show the budget--quality trade-off. The new candidate-budget comparison uses automated evaluation; the human ratings in \cref{tab:human} apply to the original evaluated videos.

\paragraph{Relation to the original retry analysis.}
Initial and Final in \cref{fig:retry-analysis} are the first and selected final outputs from the same \method{} run. The figure compares these outputs on the subset of retried shots: SF increases from .749 to .765 and Physics from 3.329 to 3.616. \Cref{app:retry-analysis} explains how this subset was selected. The new comparison above addresses the separate question of performance when baselines also receive additional generation attempts.
\FloatBarrier

\subsection{Human Audit of Verifier Reliability}
\label{app:verifier-audit}

\paragraph{Do accepted first-frame defects affect the video?}
We manually reviewed 43 L1-accepted first frames together with their resulting videos. In 3 cases (6.98\%), an initial defect persisted and affected subsequent frames; the remaining 40 were free of such inherited defects in this review. The count covers geometry, content, action-state, and composition mismatches that affect the subsequent video. It measures how often an accepted initial-state error carries over into the video.

\paragraph{Audit sample and counting unit.}
The audit covers 1,177 internal SOIG node checks associated with the 61 \method{} videos in the human-evaluation sample (37 from ViMax-Bench and 24 from ST-Bench-10) and their first frames. The audited set comprises \textbf{500 L1 first-frame checks and 677 L2 video checks}, covering object existence, spatial layout, actions, physics, factual consistency, and visible artifacts. Each item is one node's judgment in a verification record, rather than a video or a whole-shot score.

\paragraph{What counts as a verifier error?}
Three raters independently inspected the same 1,177 node checks against the corresponding requirements and media. Every item received a definite judgment. We treat a violation as the positive class: the verifier either flags a problem or accepts the check, and each rater judges whether the requirement is satisfied. Of the 1,177 checks, \method{} flagged 31 and accepted 1,146. \Cref{tab:verifier-counts} defines the four outcomes and reports their counts.

\begin{table}[!htbp]
\centering\small
\setlength{\tabcolsep}{5pt}
\caption{\textbf{Verifier decisions compared with individual human judgments.} Each row uses one rater's assessment of the same 1,177 checks as the reference.}
\label{tab:verifier-counts}
\begin{tabular}{@{}>{\raggedright\arraybackslash}p{.13\linewidth}>{\raggedright\arraybackslash}p{.19\linewidth}>{\raggedright\arraybackslash}p{.19\linewidth}>{\raggedright\arraybackslash}p{.19\linewidth}>{\raggedright\arraybackslash}p{.19\linewidth}@{}}
\toprule
 & TP & FP & FN & TN \\
Meaning & Flagged; human confirms a violation & Flagged; human finds no violation & Accepted; human finds a violation & Accepted; human confirms no violation \\
\midrule
Rater 1 & 30 & 1 & 2 & 1,144 \\
Rater 2 & 29 & 2 & 3 & 1,143 \\
Rater 3 & 27 & 4 & 5 & 1,141 \\
\bottomrule
\end{tabular}
\end{table}

\begin{table}[!htbp]
\centering\small
\setlength{\tabcolsep}{4pt}
\caption{\textbf{Four complementary error rates (\%): lower is better for every column.} Rates are calculated for each rater, then averaged. Different denominators answer different operational questions.}
\label{tab:verifier-rates}
\begin{tabular}{@{}lrrrr@{}}
\toprule
 & False discovery & False omission & False negative & False positive \\
Formula & $\frac{\mathrm{FP}}{\mathrm{TP+FP}}$ & $\frac{\mathrm{FN}}{\mathrm{TN+FN}}$ & $\frac{\mathrm{FN}}{\mathrm{TP+FN}}$ & $\frac{\mathrm{FP}}{\mathrm{TN+FP}}$ \\
\midrule
Rater 1 & 3.23 & 0.17 & 6.25 & 0.09 \\
Rater 2 & 6.45 & 0.26 & 9.38 & 0.17 \\
Rater 3 & 12.90 & 0.44 & 15.63 & 0.35 \\
\textbf{Mean} & \textbf{7.53} & \textbf{0.29} & \textbf{10.42} & \textbf{0.20} \\
\bottomrule
\end{tabular}
\end{table}
\FloatBarrier

The \textbf{false-discovery rate} is the fraction of flagged checks that are unnecessary alarms. The \textbf{false-omission rate} is the fraction of accepted checks that still contain a violation. The \textbf{false-negative rate} is the fraction of actual violations that are missed. The \textbf{false-positive rate} is the fraction of satisfied checks that are incorrectly flagged. Each rate therefore describes a specific type of verifier error with a clearly defined denominator.

These results support reliable acceptance and selective repair. On average, 99.71\% of accepted checks are confirmed as satisfied by human review; 92.47\% of flags correspond to human-confirmed violations, and 89.58\% of human-identified violations are detected. Incorrect flags affect only 0.20\% of human-satisfied checks. All averages summarize three assessments of the same 1,177 checks.

\paragraph{Residual artifacts in accepted videos.}
A fourth rater separately reviewed 50 videos accepted by the complete L2 process. Three contained visible artifacts and 47 were artifact-free, giving a residual artifact rate of 6.0\% and an artifact-free proportion of 94.0\% in this review. This video-level assessment complements the node-level error rates and the first-frame error analysis above.
\FloatBarrier

\subsection{Graph-Compilation and Verification Expenditure}
\label{app:verification-cost}

\Cref{tab:verification-cost} measures the cost of graph compilation and repeated asset, first-frame, and video verification. Together, these stages cost \textbf{\$0.1591 per shot} on average, equivalent to \textbf{7.12\% of one Seedance video-generation call} at the price used in this accounting. Graph compilation contributes \$0.0251 per shot and L0--L2 verification contributes \$0.1340. These totals include repeated checks: L1 averages 6.30 API calls per shot and L2 averages 6.68.

\begin{table}[!htbp]
\centering\small
\setlength{\tabcolsep}{5pt}
\caption{\textbf{Compilation and verification costs per shot.} The final column divides each expense by the cost of one Seedance generation. Entries are rounded independently.}
\label{tab:verification-cost}
\begin{tabular}{@{}lrrrrr@{}}
\toprule
Stage & API calls & Input tokens & Output tokens & US dollars & Relative cost \\
\midrule
Graph compilation & 2.13 & 12,291 & 8,010 & 0.0251 & 1.12\% \\
L0: portrait verification & 1.19 & 5,662 & 1,592 & 0.0065 & 0.29\% \\
L1: first-frame verification & 6.30 & 58,960 & 12,650 & 0.0586 & 2.62\% \\
L2: video verification & 6.68 & 94,644 & 9,782 & 0.0689 & 3.08\% \\
\midrule
\textbf{Total} & \textbf{16.30} & \textbf{171,558} & \textbf{32,035} & \textbf{0.1591} & \textbf{7.12\%} \\
\bottomrule
\end{tabular}
\end{table}

Graph compilation and repeated verification thus add a modest expense relative to a single video-generation call. This finding complements the 24--25\% reduction in generation-plus-edit operations in \cref{tab:ablation} and the 52.7\% reduction in generation attempts in \cref{tab:three-candidate}. The table measures the four listed stages; a full pipeline cost would also include image synthesis, video generation/editing, and prompt refinement. Operation counts and these measured verification expenses describe complementary aspects of efficiency.
\FloatBarrier

\subsection{Isolating the Effect of Dependency Edges}
\label{app:edge-replays}

In \cref{fig:pipeline}, category branches group related checks, dependency edges specify prerequisites, and each judgment links to its visual evidence. To isolate the contribution of dependency edges, we rerun scoring and repair-target selection on existing verification results. The video, atomic checks, answers, and observations stay fixed; only the edges change. We refer to this controlled recalculation as a replay.

\paragraph{Preventing unsupported downstream credit.}
In the board-game cafe example \texttt{scene\_1\_shot\_2}, candidate \texttt{2:video:0}, action check A1 fails: the requested card-dealing action is absent and the character holds a phone instead. P1 nevertheless answers yes to valid hand--object contact for the specified actions. The graph prevents this downstream answer from contributing as though its prerequisite action had occurred.

\begin{table}[!htbp]
\centering\small
\caption{\textbf{Dependency edges change which checks receive credit.} All 11 raw node judgments remain fixed. Recalculation with the original edges reproduces the score from the run.}
\label{tab:edge-credit}
\begin{tabular}{@{}lrl@{}}
\toprule
Edges used & Credited checks & Consequence \\
\midrule
Original graph & $6/11$ & Failed prerequisites block downstream credit \\
Only A1 $\rightarrow$ P1 removed & $8/11$ & P1 and its descendant F1 regain credit \\
All dependency edges removed & $10/11$ & Each raw yes is credited independently \\
\bottomrule
\end{tabular}
\end{table}

Removing a single edge therefore changes the diagnosis: checks whose prerequisites failed now receive credit. The fraction in \cref{tab:edge-credit} measures credited requirements in the verifier. A1 remains a critical failure in all three settings, and the candidate is rejected in each case.

\paragraph{Prioritizing the missing prerequisites.}
In a second board-game cafe shot, \texttt{scene\_1\_shot\_3}, candidate \texttt{3:video:0}, four checks fail. A1 reports that the hourglass is not flipped during 5--7 seconds; A2 reports that the card is placed before, rather than within, that interval. P1 flags the interactions involving these missing or mistimed actions. P2 reports no hourglass flip with which to assess the required sand flow.

\begin{table}[!htbp]
\centering\small
\caption{\textbf{Dependency edges distinguish primary repair targets from dependent failures.} All four failed requirements remain part of the repair task.}
\label{tab:edge-priority}
\begin{tabular}{@{}>{\raggedright\arraybackslash}p{.21\linewidth}>{\raggedright\arraybackslash}p{.30\linewidth}>{\raggedright\arraybackslash}p{.41\linewidth}@{}}
\toprule
Configuration & Primary repair targets & Dependent failed requirements \\
\midrule
With dependencies & A1: flip the hourglass; A2: correct card-placement timing & P1: contact during these actions; P2: sand flow after the flip \\
Without dependencies & A1, A2, P1, and P2 at the same priority & No dependency-based separation \\
\bottomrule
\end{tabular}
\end{table}

With dependencies, repair-target selection prioritizes the missing hourglass flip and the incorrect card-placement timing. The contact and sand-flow requirements remain explicit dependent targets. Removing the edges instead presents all four failures at equal priority. This comparison isolates how the graph changes repair instructions by rerunning the selection code on the same observations. It evaluates that decision step; generation quality is assessed separately below.

\paragraph{A corresponding failure in generated outputs.}
The AeroPress example (barista coffee cultures, \texttt{scene\_2\_shot\_5}) illustrates why a prerequisite can matter even when global Physics scores agree. The external evaluator observes \method{} pouring water into the brewer at 2.09--3.43 seconds and pressing at 4.57--5.71 seconds, and marks event order PASS. Flat \texttt{abl1} pours directly into the mug and then presses the brewer; Flat \texttt{abl2} omits pressing before drinking. Both receive event-order FAIL, although \method{} and \texttt{abl1} both receive global Physics 4. The missing prerequisite is specific: pressing must act on water previously added to the brewer. This comparison of complete configurations illustrates the failure mode; the controlled replays above isolate the edge-dependent decisions.
\FloatBarrier

\subsection{From a Failed Check to a Verified Repair}
\label{app:repair-evidence}

\paragraph{Inputs, outputs, and update rules.}
The refiner receives the original requirements, current generation prompt, failed image or video, and observations for the failed checks. It returns revised prompt text and a structured list identifying each failed node, the field to edit, the proposed change, and its rationale. Before applying an edit, the program checks that it refers to an eligible target and field, supplies usable text, and changes the current draft. The original requirements and all other fields stay fixed. L1 edits the image description; L2 edits the video prompt while keeping the selected first frame. Each new candidate is then verified against the original requirements.

\begin{table}[!htbp]
\centering\small
\caption{\textbf{Acceptance follows explicit validity and rejection checks.} The rules are applied in order; a score threshold is used only for valid older-format reports that lack an explicit decision.}
\label{tab:acceptance-rules}
\begin{tabular}{@{}>{\raggedright\arraybackslash}p{.33\linewidth}>{\raggedright\arraybackslash}p{.62\linewidth}@{}}
\toprule
Report condition & Decision \\
\midrule
Invalid or unscored report & Require a valid, scored report before accepting the candidate. \\
Hard rejection or veto & Reject the candidate regardless of its graph-coverage score. \\
Valid report with explicit judge decision & Use that decision after the hard-rejection check. \\
Valid legacy report without an explicit decision & Compare its finite score with the configured threshold. \\
Budget exhaustion & Return a permitted fallback with its original verdict and the reason for termination. \\
\bottomrule
\end{tabular}
\end{table}

First-frame verification checks identity, object inventory, static layout, and physical support. Video verification checks requirements that depend on motion or change over time. The optional relaxed video policy can waive secondary environmental effects. Each waiver is identified as a policy adjustment, and the original visual observation remains available alongside the adjusted verdict and final acceptance decision.

\paragraph{An executed cooking-scene repair.}
The cooking run in \cref{fig:teaser} shows the full repair process. The initial video report identifies food appearing on the right plate during 4--6 seconds without visible placement, along with a clock-hand discontinuity and anatomical/clothing defects. Both required-action checks A1 and A2 are satisfied. The graph fraction is $10/11=.9091$, giving a base score of $1+9(10/11)=9.18$, above the configured threshold of 7.0. The artifact rejection rule then reduces the final score to 3.0, and the candidate fails.

The refiner revises the motion field to distinguish Chef A's vegetable placement from Chef B's sauce and garnish actions, specify smooth timer behavior, and maintain coherent visible anatomy and clothing. The next generation request uses this revised text verbatim and the same first-frame reference. The subsequent report uses the same contract ID, \texttt{soig-10-0009}, and marks the artifact check, A1, and A2 as satisfied, with a final PASS decision. This example connects each step: \emph{observed failure $\rightarrow$ prompt revision $\rightarrow$ generation request $\rightarrow$ verification against the same requirements}.

The rules above describe the implementation examined for this revision. The cooking example comes from an earlier run, whose prompt wording may differ. Its execution log confirms the specific edit, generation request, and verification outcome described here.
\FloatBarrier

\subsection{Human-Study Workload and Matched-Video Disagreements}
\label{app:human-workload}

The 768 and 2,322 human ratings count valid assessments of individual system--shot videos. A video can receive ratings from several participants. Within each benchmark, every comparison group presents one video from each method. The total counts follow from clip duration, groups per participant, participant counts, and invalid-score exclusion.

\begin{table}[!htbp]
\centering\small
\caption{\textbf{Human-study accounting.} Viewing duration sums all four system videos per assigned group and excludes reading, replaying, deliberation, and data entry.}
\label{tab:human-workload}
\begin{tabular}{@{}lrr@{}}
\toprule
Quantity & ViMax-Bench & ST-Bench-10 \\
\midrule
Clip duration & 15 s & 5 s \\
Participants & 16 & 25 \\
Shot groups per participant & 12 & 24 \\
System videos per group & 4 & 4 \\
Single-play video duration per participant & 720 s & 480 s \\
Distinct groups across the study & 37 & 24 \\
Distinct videos across the study & 148 & 96 \\
Submitted rating records & 768 & 2,400 \\
Excluded out-of-range records & 0 & 78 \\
Valid rating records & 768 & 2,322 \\
\bottomrule
\end{tabular}
\end{table}

The studies aimed for comparable overall assessment workload. ST-Bench-10 assigns more, shorter groups; ViMax-Bench assigns fewer, longer groups. In practice, overall assessment times were comparable after including requirement reading, video inspection, deliberation, and rating entry. The table lists single-play video duration; these additional activities account for the rest of each session. Automated results are reported separately for the 35 ViMax-Bench cases and the ten ST-Bench-10 cases.

\paragraph{Understanding human and automated score differences.}
The original human and automated tables summarize different video samples with different aggregation weights. Evaluator agreement is assessed by comparing scores on the same videos. For this analysis, we average the human ratings for each system--shot video and pair that mean with its automated score, as described in \cref{app:human}. System rankings summarize overall performance, while video-level correlations measure agreement across individual outputs.

The rating records also show defects that the automated evaluator missed. In all three examples in \cref{tab:human-disagreements}, it assigns global Physics 5 and reports no artifact events, while human comments identify a visible error. These cases show how human inspection detects specific errors behind score disagreements. They illustrate failure types; the systematic verifier error-rate assessment is reported in \cref{app:verifier-audit}.

\begin{table}[!htbp]
\centering\small
\setlength{\tabcolsep}{4pt}
\caption{\textbf{Same-video errors identified in human comments.} Human means use all valid ratings for each listed video.}
\label{tab:human-disagreements}
\begin{tabular}{@{}>{\raggedright\arraybackslash}p{.21\linewidth}>{\raggedright\arraybackslash}p{.14\linewidth}>{\raggedright\arraybackslash}p{.37\linewidth}rr@{}}
\toprule
System / case & Shot & Human-observed issue & Ratings & Human Phys \\
\midrule
AniMaker / \texttt{phd} & \texttt{scene\_0}\newline\texttt{shot\_0} & Code displayed on the back of the laptop lid & 23 & 2.522 \\
AniMaker / \texttt{japan} & \texttt{scene\_7}\newline\texttt{shot\_7} & Umbrella shaft moves into the hand without a plausible grasp & 25 & 2.520 \\
MovieAgent / \texttt{black} & \texttt{scene\_1}\newline\texttt{shot\_1} & Hand--guitar and case--guitar interpenetration during removal & 25 & 2.560 \\
\bottomrule
\end{tabular}
\end{table}

The original blinded human evaluation ranks \method{} first in both mean SF and mean Physics on each benchmark sample. The cases above assess the external evaluator. The node audit in \cref{app:verifier-audit} separately measures false-positive and false-negative rates for the internal verifier.
\FloatBarrier

\subsection{Cross-Shot Character Continuity}
\label{app:cross-shot}

\method{} supports continuity through shared character definitions and reusable reference assets. Each shot's stage-specific requirements also stay fixed during repair. In the mentor--student woodworking case, the story moves from instruction in \texttt{scene\_1\_shot\_4} to the same student operating the saw in \texttt{scene\_2\_shot\_5}.

The \method{} generation requests use the same student reference image in both shots, confirmed by the identical SHA-256 prefix \texttt{c48f946790c85dfe}. Both L1 reports also check the student's appearance against the original requirements. In the compared ViMax output, the student changes across the cut from a blond, clean-shaven person in a navy shirt to a dark-haired, bearded person in a light shirt and dark apron. \method{} preserves the student's principal identity traits across the same transition, as shown in \cref{fig:cross-shot-evidence}.

\begin{figure}[!htbp]
\centering
\begin{tabular}{@{}c@{\hspace{6pt}}c@{}}
\multicolumn{1}{c}{Earlier shot: 14.922 s} & \multicolumn{1}{c}{Next shot: 0.000 s} \\
\includegraphics[width=.48\linewidth]{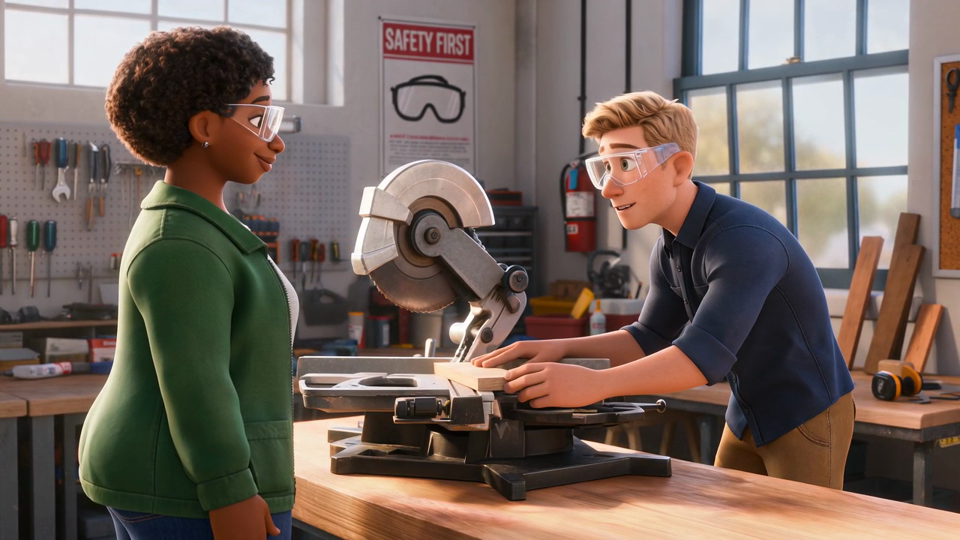} &
\includegraphics[width=.48\linewidth]{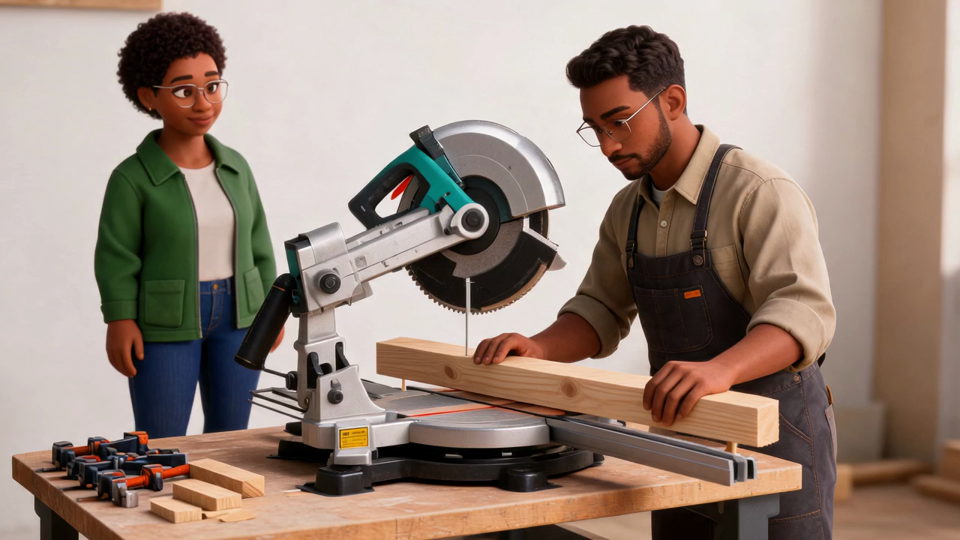} \\
\multicolumn{2}{c}{ViMax: student's hair, facial hair, and clothing change across the cut} \\[6pt]
\includegraphics[width=.48\linewidth]{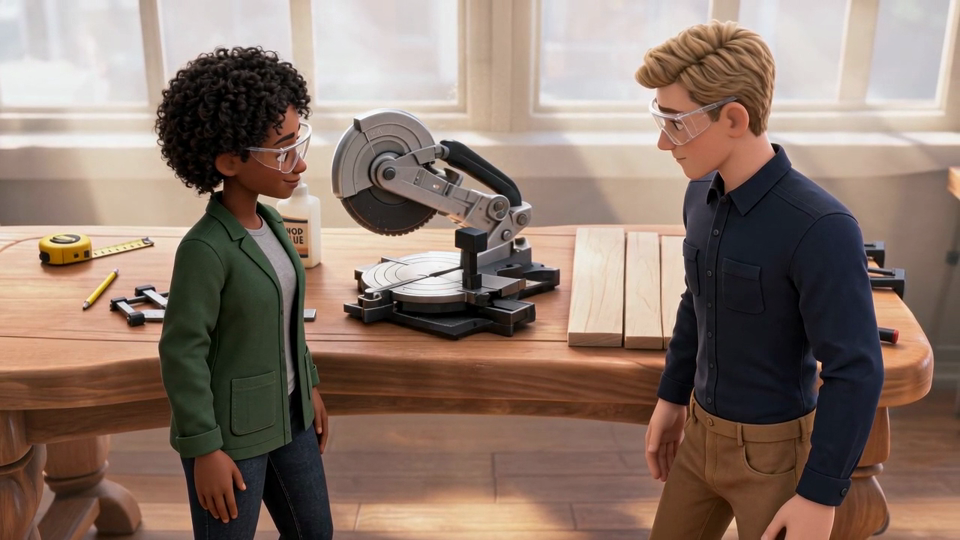} &
\includegraphics[width=.48\linewidth]{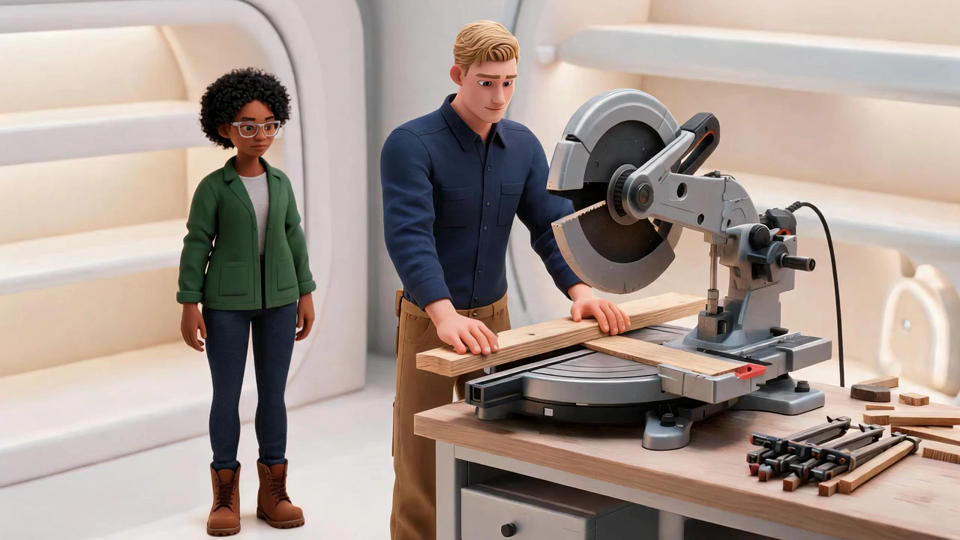} \\
\multicolumn{2}{c}{\method{}: blond hair, clean-shaven appearance, and navy shirt retained}
\end{tabular}
\caption{\textbf{Cross-shot identity continuity in the woodworking example.} Rows compare consecutive outputs of each evaluated end-to-end pipeline. Frames are unretouched and retain their full field of view. The corresponding source story requests the same student on both sides of the cut.}
\label{fig:cross-shot-evidence}
\end{figure}

This example demonstrates character continuity across a shot transition, supported by the generated frames and the shared reference image. It complements \cref{sec:limitations}: stable references already help preserve identity, while explicit checks of state transitions and systematic full-story evaluation are further capabilities to develop.
\FloatBarrier
\endgroup

%% file: appendix_refiner_prompts.tex
\clearpage
\begingroup

\raggedbottom
\captionsetup{font={small,color=black},labelfont={bf,color=black},textfont={color=black}}
\newtcblisting{giverrefinerbox}[1]{
  enhanced,breakable,listing only,listing engine=listings,
  title={#1},fonttitle=\small\bfseries,
  title after break={#1\enspace (continued)},
  colback=giverPromptBlue!2!white,colframe=giverPromptBlue!70!black,
  colbacktitle=giverPromptBlue!9!white,coltitle=black,colupper=black,
  boxrule=0.45pt,arc=1pt,outer arc=1pt,
  left=6pt,right=6pt,top=5pt,bottom=5pt,
  before skip=7pt,after skip=9pt,pad at break*=2pt,
  listing options={basicstyle=\fontencoding{T1}\ttfamily\fontsize{8}{9.6}\selectfont,
    columns=fullflexible,keepspaces=true,breaklines=true,breakatwhitespace=false,
    breakindent=0pt,showstringspaces=false,showspaces=false,showtabs=false,
    numbers=none,tabsize=2,aboveskip=0pt,belowskip=0pt,upquote=true}
}
\section{Refiner Prompts and Field-Update Rules}
\label{app:refiner-prompts}
When a check fails, GIVER gives a repair MLLM the original requirements,
current prompt, failed image or video, and the verifier's observations.
The MLLM proposes an updated generation prompt and identifies the failed
checks addressed by each edit.
The program validates the edited fields, then generates a new candidate and
checks it against the same requirements.
In the reported experiments, L1 revised the first-frame description and L2
revised the video prompt while keeping the selected first-frame image fixed.

The complete templates below instruct the repair MLLM; its output supplies
the revised text for the image or video generator. Shared instructions appear
in full in both templates, and braces mark values filled in for each call.
\textbf{The optional L2-to-L1 return path in the general interface was disabled
in all reported experiments.} \Cref{app:repair-evidence} explains the version
scope of the implementation and cooking example. Matching text files and
source excerpts are in the supplementary \texttt{repair/} directory.

\subsection{Messages and Template Inputs}
The system message contains the L1 or L2 template with the inputs in
\cref{tab:refiner-inputs} filled in. For an ST-Bench shot that continues from
the preceding shot's last frame, the system appends instruction L2-C to L2.
The exact condition is \nolinkurl{resolved_mode=previous_last_frame},
\nolinkurl{generation_hints.source=st_bench_adapter}, and
\nolinkurl{generation_hints.resolved_cut=false}.
The user-role message contains a JSON text block with \texttt{contract\_id},
followed by the media attachments. L1 also supplies \texttt{shot\_references},
the failed first frame, and available character and scene reference images.
L2 supplies the complete failed video and its selected first frame.

The L1 candidate-image label is ``Current failed first-frame evidence, not the
target: repair the evidenced failed relationships; preserve only correct
details.'' The L2 first-frame label is ``Current first-frame candidate; preserve
correct identity, props and geometry.'' The message includes the images or
video themselves as attachments.

\begin{table}[!htbp]
\centering\small
\renewcommand{\arraystretch}{1.12}
\caption{\textbf{Inputs supplied to the repair MLLM.} Each placeholder is filled with the corresponding prompt, fixed requirement, or verification evidence.}
\label{tab:refiner-inputs}
\begin{tabular}{@{}>{\raggedright\arraybackslash}p{0.32\linewidth}>{\raggedright\arraybackslash}p{0.64\linewidth}@{}}
\toprule
\textbf{Template input} & \textbf{Binding} \\
\midrule
\texttt{current\_generation\_draft} & Current \texttt{ff\_desc}, falling back to \texttt{visual\_desc}. \\
\texttt{visual\_desc}, \texttt{motion\_desc} & Current generation fields, with authored shot descriptions as fallbacks. \\
\texttt{repair\_requirements} & Selected fixed targets: \texttt{node\_id}, \texttt{source\_node\_id}, \texttt{must\_hold} (or description), \texttt{question}, and \texttt{entities}. Filter for nodes whose \texttt{match} differs from true; use the selected nodes if that filter yields no IDs. \\
\texttt{repair\_targets} & The same targets, with their \texttt{observed} and \texttt{match} values from verification. \\
\texttt{repair\_diagnosis} & \texttt{interaction\_checks}: \texttt{node\_id}, \texttt{interaction\_id}, \texttt{observed}, \texttt{match}, and \texttt{severity}; plus \texttt{penetration\_detected} and \texttt{penetration\_detail}. \\
\texttt{reference\_context} & \texttt{authored\_reference}, the last two \texttt{output\_correction\_feedback} entries, and \texttt{last\_unaccepted\_draft} when present. The original requirements remain fixed across revisions. \\
\texttt{shot\_duration\_seconds} & Read-only duration from the frozen contract. \\
\bottomrule
\end{tabular}
\end{table}

The diagnosis copies each verification result's \texttt{satisfied} value to
\texttt{match}. Severity is critical for explicitly critical nodes, or hard
nodes whose critical flag is unset. The penetration fields start as
\texttt{false} and an empty string; the MLLM also inspects the attached media.
The shot duration remains fixed, and the repair inputs exclude verification
sampling schedules.
\FloatBarrier

\subsection{Complete L1 and L2 Templates}
Each repair call uses the complete template for its stage. For the continuation
condition above, L2-C follows the complete L2 template as an additional instruction.
\begin{giverrefinerbox}{L1: First-frame refinement}
[System instruction]
You write first-frame image-editing instructions. Inspect the attached failed candidate directly. The current prompt is the plan that produced the erroneous image, not a target that must be retained.
[Current generation draft] {current_generation_draft}
[Itemized repair basis: frozen targets and observations of this image] {repair_targets}
[Original shot reference and format-correction feedback] {reference_context}
Do not let prompt revisions go out of shot description: rewrite locally only to satisfy existing failed nodes. Do not add actions, persistent states, camera movements, or restrictions that the original shot description / authored_visual / authored_motion does not require. For example, "pause briefly before touching" must not become "remain motionless throughout."
[Boundaries of the source shot]
The original visual description, setting, characters, and explicit actions jointly define this shot's scope. Background narrative, prior events, internal thoughts, identity, and changes across days provide context; they do not automatically become additional visible events required in this shot.
Allow the handling, contact, and natural reactions necessary to complete the original action. Do not invent badges, notices, genuine comparison objects, screen captions, additional people, abrupt weather changes, or within-shot transitions to prove abstract story content. Preserve such content when explicitly required. This is not a keyword blacklist.
Depicting "becoming the leader" as a falling badge, "identifying a counterfeit" as another genuine object on the table, or a cooling-weather background as accumulated snow within seconds is not necessary action elaboration. Do not assign existing objects new story meanings such as genuine artifacts or evidence without authorization.
Extract only propositions grounded in the source and checkable in the current medium into the SOIG. Do not force pixel evidence for every background statement. Retain general identity, count, and physical-continuity checks. caption_hint, beat_summary, and dependency edges must not introduce new requirements; a dependency denotes a genuinely necessary prerequisite.
Explicitly identify contradictions or insufficient information in the source rather than inventing a story to resolve them. The original storyboard takes priority over model-generated graph summaries, incorrectly generated initial-frame content, extracted_actions, revision suggestions, and the current prompt.
Before responding, check within this reasoning step that each added entity, event, attribute, and scene change is supported by the source or necessary to connect the original action. Remove unsupported additions without adding output fields, checking nodes, or model calls.
[Scope of interaction requirements]
The number of character instances is not the number of visible hands. A singular "hand" in the source describes a body part involved in an action; it does not automatically mean that only one hand may appear.
Using two tools does not mean holding both in the same hand simultaneously. Unless the source explicitly specifies the same hand, a left or right hand, or simultaneous holding, allow natural allocation and handling consistent with the action. Do not freeze one generation plan as the only valid implementation.
When the source merely lists multiple tools, existing nodes should check tool identity, natural support and accessibility, and actual contact. Do not use a singular-subject conjunction such as "the hand contacts A and B" to force one hand to contact both tools simultaneously. Do not invent a fixed allocation when the source does not specify one. Retain a same-hand simultaneous-operation condition only when the source explicitly requires it.
Preserve each tool's actual purpose and structure. Do not replace a palette knife or table knife with an unspecified object that "looks like a knife." Arrange props so they can be naturally reached and supported; do not force every tool into a hand at once.
These boundaries apply to question, must_hold, caption_hint, beat_summary, action_readiness, and generation or revision text. Preserve explicit one-hand or two-hand constraints and do not alter the original action.

Change only the visual relationships associated with failed nodes and keep other correct sentences verbatim. Fixed node requirements are the only repair targets; observations are evidence, not new requirements.
For local_edit, replace the corresponding original sentence with an explicit image-editing operation: identify the erroneous object/state and specify what to add, remove, or change it to. Even if the original sentence states the target but the image fails to realize it, rewrite it as an actionable edit; do not merely repeat it or add "clearly," "completely," "100%," or "prominently."
Keep framing and visibility consistent throughout the complete draft. A waist-up view or face/hand close-up does not additionally require feet, legs, or the floor. Walking can continue in that framing; do not switch to a full-body view merely to show a step. When repairing framing, foot/full-body descriptions that incorrectly widen the view belong to the same failed relationship; replace them with wording consistent with the original framing rather than retaining contradictions.
Repair scale/perspective only for relationships that are actually visible and demonstrably wrong. A missing applicable reference does not prove a scale error and does not justify changing clothing, expression, or lighting. Do not add people or a workbench to an empty scene. Do not expand an absence-of-people requirement into prohibitions on windows, pillars, or other setting elements, or invent subject-area percentages or a single-light-source standard.
For example, if a character is missing, write "add the specified character in the specified clothing at the originally required position while retaining the existing correct characters," not merely "the character is clearly visible." Replace incorrect clothing with the required clothing. These examples illustrate wording; they are not new content for this shot.
First resolve character identity and count, necessary props, action-compatible positions, and supported contact. Replace erroneous sentences that conflict with these targets; do not retain a wrong pose and append the correct pose. The first frame is a static state and need not show an action already completed.
When repairing an existing prop, change its state directly while preserving its instance, position, and count. Do not add a correct substitute while leaving the erroneous original instance.
Use the attached character identifiers to bind identity, clothing, pose, position, and contacted objects to the same instance within that character's original sentence. Do not assign one person's clothing or action to another. A reference image does not imply an extra person in the scene. Base clothing on the frozen requirements and that character's reference; base position on the original shot composition and frozen nodes. Do not copy a portrait pose or empty-scene framing. Inspect the current image, character references, and scene reference before deciding the edit scope, not the case name.
Preserve original requirements and correct content, not erroneous counts, clothing, or positions. Normal occlusion is not a defect. Do not add unobstructed visibility, full-body framing, exact distances, or unsupported spatial constraints. Unknown/unobservable is not a confirmed defect.
Choose anchor_repair_mode as local_edit or full_regen from the actual image. Choose local_edit if a local edit can accommodate the required people and props. Choose full_regen if the current framing/layout cannot accommodate the original requirements, implementing the original composition without inheriting the erroneous base image. Do not select regeneration solely from the node category or round number.
For full_regen, the complete draft describes only the target image, without base-image operations such as "delete the garbled text," "move the original object away," or "replace the current incorrect clothes." Use such operations only for local_edit, retaining only those needed in the current round and replacing completed or irrelevant operations in their corresponding sentences.
When repairing screens, signs, or chat text, preserve explicitly specified short names/text and the original meaning. Do not invent longer dialogue, additional names, or emoji. Make the original display region's framing, perspective, and typography sufficient to read the necessary content; do not add a second screen, subtitles, or new text props. Do not simply repeat "clear" or "readable." Keep unspecified verbatim wording brief and natural, and do not turn a current example sentence into an immutable future requirement.
Place edits in the failed object's original sentence and still output one complete revised draft, ff_desc_revised. Do not append a patch paragraph or merge historical revisions. No verbatim quotation or source_span_id is needed. The runtime uses the complete draft directly rather than concatenating critical_fixes.
critical_fixes briefly records actual changes in this draft: node_id, target_field=visual, proposed_fix, and rationale. changes_summary describes only what the draft says; it must not claim that an image not yet generated is already repaired.
If last_unaccepted_draft exists, correct the format feedback and retain its correct changes, without inheriting added constraints. Before responding, confirm that the actual correction appears in ff_desc_revised, not only in the change record.
Keep identity, style, unrelated camera viewpoint, and original timing unchanged. The duration, {shot_duration_seconds} seconds, is read-only. Do not add diagnosis sampling timestamps. Do not truncate correct content or compress it to meet a word count.
Leave visual_desc_revised and motion_desc_revised empty, and node_caption_updates an empty object. Return JSON only.
\end{giverrefinerbox}
\begin{giverrefinerbox}{L2: Video refinement}
[System instruction]
You are a video-prompt refiner. Output a complete revised draft, making only small changes to descriptions related to the SOIG nodes that failed in this round.
[Current first-frame visual] {visual_desc}
[Current video-generation prompt] {motion_desc}
[Frozen requirements of failed nodes, excluding verification sampling timestamps] {repair_requirements}
[Current video diagnosis: observation descriptions are not new generation requirements] {repair_diagnosis}
[Original shot reference and output-correction feedback, not historical repair requirements] {reference_context}
Do not let prompt revisions go out of shot description: rewrite locally only to satisfy existing failed nodes. Do not add actions, persistent states, camera movements, or restrictions that the original shot description / authored_visual / authored_motion does not require. For example, "pause briefly before touching" must not become "remain motionless throughout."
[Boundaries of the source shot]
The original visual description, setting, characters, and explicit actions jointly define this shot's scope. Background narrative, prior events, internal thoughts, identity, and changes across days provide context; they do not automatically become additional visible events required in this shot.
Allow the handling, contact, and natural reactions necessary to complete the original action. Do not invent badges, notices, genuine comparison objects, screen captions, additional people, abrupt weather changes, or within-shot transitions to prove abstract story content. Preserve such content when explicitly required. This is not a keyword blacklist.
Depicting "becoming the leader" as a falling badge, "identifying a counterfeit" as another genuine object on the table, or a cooling-weather background as accumulated snow within seconds is not necessary action elaboration. Do not assign existing objects new story meanings such as genuine artifacts or evidence without authorization.
Extract only propositions grounded in the source and checkable in the current medium into the SOIG. Do not force pixel evidence for every background statement. Retain general identity, count, and physical-continuity checks. caption_hint, beat_summary, and dependency edges must not introduce new requirements; a dependency denotes a genuinely necessary prerequisite.
Explicitly identify contradictions or insufficient information in the source rather than inventing a story to resolve them. The original storyboard takes priority over model-generated graph summaries, incorrectly generated initial-frame content, extracted_actions, revision suggestions, and the current prompt.
Before responding, check within this reasoning step that each added entity, event, attribute, and scene change is supported by the source or necessary to connect the original action. Remove unsupported additions without adding output fields, checking nodes, or model calls.
[Scope of interaction requirements]
The number of character instances is not the number of visible hands. A singular "hand" in the source describes a body part involved in an action; it does not automatically mean that only one hand may appear.
Using two tools does not mean holding both in the same hand simultaneously. Unless the source explicitly specifies the same hand, a left or right hand, or simultaneous holding, allow natural allocation and handling consistent with the action. Do not freeze one generation plan as the only valid implementation.
When the source merely lists multiple tools, existing nodes should check tool identity, natural support and accessibility, and actual contact. Do not use a singular-subject conjunction such as "the hand contacts A and B" to force one hand to contact both tools simultaneously. Do not invent a fixed allocation when the source does not specify one. Retain a same-hand simultaneous-operation condition only when the source explicitly requires it.
Preserve each tool's actual purpose and structure. Do not replace a palette knife or table knife with an unspecified object that "looks like a knife." Arrange props so they can be naturally reached and supported; do not force every tool into a hand at once.
These boundaries apply to question, must_hold, caption_hint, beat_summary, action_readiness, and generation or revision text. Preserve explicit one-hand or two-hand constraints and do not alter the original action.
motion_desc_revised is the complete video prompt ready for submission. Preserve the current prompt's language, organization, and all timestamps.
Only when the defect is in the first frame, also output a complete first-frame revision in visual_desc_revised. Unmodified fields may be empty; the runtime retains their current text.
critical_fixes records only node_id, target_field (visual or motion), proposed_fix (the new relevant description), and rationale.
There is no need to quote the original sentence verbatim or provide source_span_id. The runtime uses the complete revision rather than concatenating change records.
Fixed node requirements are the only repair targets. Diagnoses explain where the current pixels fail to satisfy them; they are not new requirements.
Distinguish the initial state from the action: change the first frame only for evidenced initial-state defects. A later video failure does not automatically require a raised foot, open mouth, or completed action in the first frame.
When the action can naturally begin from a still pose, do not invent a suspended foot, raised hand, or speaking mouth shape as "preparation."
The current prompt is a generation plan to revise, not the contract. In particular, do not automatically retain restrictions from previous revisions in sentences associated with failed nodes.
Compare the fixed requirements and original shot reference, then re-express the failed relationship. Replace only the relevant sentences and leave other correct content unchanged.
Clarify reasonable relative depth, orientation, and contact, but do not add unsupported visibility, complete occlusion, left/right exclusion, or exact-position restrictions.
When the same relationship fails again, change its implementation instead of escalating "completely," "100%," or "never" each round. Directly replace excessive restrictions in the old sentence.
For local image edits, identify the erroneous object in the base image, its current state, and its target state. For example, change the two white pillars just right of center to matte black while keeping their positions and shapes, rather than merely changing "black" to "solid/opaque black."
Place the explicit edit in the corresponding object's original sentence, not in an appended repair paragraph. Still output one complete prompt, keeping other correct sentences unchanged.
The image generator receives only images and the final prompt. It does not know what "failed node," "frozen SOIG," or "specified location" refers to. Use concrete visual language rather than these internal terms as substitutes for the target.
"Hover" requires a visible gap, not an invented 2-3 cm distance. "Dominates foreground right" does not prohibit the counter from appearing anywhere on the left. Do not add unsupported dimensions, full-visibility conditions, or exclusions.
Inspect the failed candidate pixels attached to the request directly; do not imagine an image from the diagnosis. Identify the smallest erroneous relationship and preserve the layout that is already correct.
Replace the entire sentence describing that relationship rather than appending a prohibition. Remove restrictions introduced by an earlier revision but absent from the frozen requirements.
For example, "the hand is behind the globe toward the cabinet-side seam" does not require the entire hand to be hidden or prohibit its appearance to the right of the globe. Do not change the original camera viewpoint.
Do not repeat historical patches or the whole contract, and do not append a repair paragraph. Leave node_caption_updates empty; revisions must not change the verification target.
Before output, check the complete revision internally without adding output fields:
- When repairing an existing prop or display, change the erroneous state of the original instance rather than introducing a replacement. First identify its position, appearance, and count.
  Do not add a correct object to satisfy a node while leaving the erroneous original object in the image. Add an instance only when the frozen requirements explicitly establish that an instance is missing.
- last_unaccepted_draft is the last complete draft rejected by format/interface checks, not the contract or an accepted result. If present, use it as the correction starting point, fix its errors against the frozen requirements, and retain its correct changes. Do not restart from the original or blindly retain stricter constraints introduced by the draft. Carry only this last draft, not a combination of all earlier patches.
- critical_fixes must list every modified field retained in this complete revision relative to the current generation draft, including correct visual changes from the previous draft.
- First-frame changes such as character count and initial prop position must remain in the first-frame visual field; describing them only in motion cannot change an existing first frame.
- Check that repair sentences do not contradict actions in the same draft. Preventing a prop from disappearing does not prohibit picking it up; waist height does not mean the object's top edge may never exceed waist height. Visibility during a specified action does not imply visibility throughout. Do not compensate for unverifiable off-screen content or missing verification references by changing clothes, adding character features, or changing the camera viewpoint.
- Check that prop origins, character roles, and relative positions related to the failed node remain consistent with the frozen requirements; do not inherit errors in the current prompt. Preserve unrelated correct sentences and timing. If the existing timing genuinely conflicts with a frozen requirement, correct only that failed node's explicit conflict rather than rescheduling the whole clip.
Normal synonymous wording does not require verbatim reproduction of proposed_fix. Do not repeat sentences to satisfy string matching.
Write directly executable visual instructions for the generator, not an inspection report:
- Identify characters through clothing, position, and role visible in the current reference or first frame; use names only as additional identifiers. With multiple people, do not use "one of them" instead of specifying the actor.
- For close-ups, describe only the hands, cuffs, and prop ownership actually in frame. Do not require faces or full bodies to enter the frame solely for identity verification. Preserve identity references.
- Rewrite only the complete sentence for the failed relationship, specifying which erroneous object changes to which correct state. Do not keep the old sentence and pile on synonymous negations or progressively add "completely," "never," or "no sideways displacement."
- For count errors, specify which people to retain and which extra instance to remove. Identify each prop's holder and contact point. Do not guess the location of an extra instance without pixel evidence.
- State the style clearly and concisely once, preserving the original requirement without repeated rendering adjectives. Keep unrelated correct sentences verbatim and do not truncate them mechanically.
- Do not turn unobservability, format errors, or sampling timestamps into new generation requirements. Do not add unsupported poses, mouth shapes, occlusion ratios, or exact distances.
Action performers must correspond to people visible in the first frame. Disambiguate them by clothing, position, and role. Distinguish fixed equipment housings from movable parts.
Start from actual holding and contact states in the first frame. Specify handling, motion paths, contacted objects, and final states without redesigning the initial composition.
Keep other correct original sentences, identity, style, camera viewpoint, action order, and original time descriptions unchanged. Earlier character quotas do not apply.
The shot duration, {shot_duration_seconds} seconds, is read-only. Do not add schedules or pauses, or insert diagnosis sampling timestamps into the description.
Do not put inspection explanations into proposed_fix. Retain original timestamps; do not schedule extra actions or pauses to make verification easier.
Output node_caption_updates as an empty object and summarize modified nodes in changes_summary. Return JSON only.
\end{giverrefinerbox}
\begin{giverrefinerbox}{L2-C: Conditional continuation instruction}
[Conditional system suffix]
ST-Bench cut=false continuation: the attached predecessor last frame is the immutable t=0 source. The standalone first_frame_prompt is NOT a second t=0 target. Inspect initial continuity against the attached frame; use the authored video description for subsequent action and framing. Allow only physically continuous movement/reframing necessary for that action, never an instant reset, teleportation, invented prop or extra story event. Keep identity, contact, geometry and causal checks. If the authored action truly cannot proceed from this frame, report route_storyboard_conflict with evidence; do not excuse it as a pass. For repairs preserve visual_desc byte-for-byte and target only motion. Never request an image edit, removal of an authored action, or an opening-state replacement.
\end{giverrefinerbox}

\clearpage
\subsection{Output Fields and Update Rules}
\textbf{Node-linked output.} Each \texttt{critical\_fixes} entry identifies
\texttt{node\_id}, \texttt{target\_field} (\texttt{visual} or \texttt{motion}),
\texttt{proposed\_fix}, and \texttt{rationale}. The node ID and proposed fix
must be nonempty. Complete-draft replacement accepts an omitted
\texttt{original\_snippet}. Responses must use only the declared fields.

\textbf{L1.} The response returns \texttt{ff\_desc\_revised} and
\texttt{anchor\_repair\_mode} (\texttt{local\_edit} or \texttt{full\_regen};
an omitted mode defaults to \texttt{local\_edit}).
The L1 prompt leaves \texttt{visual\_desc\_revised} and
\texttt{motion\_desc\_revised} empty. \textbf{L2.} The response returns
\texttt{motion\_desc\_revised}. In all reported experiments, L2 kept the
selected first-frame image fixed. The general interface also supports
\texttt{visual\_desc\_revised} for an initial-image defect and a return to L1;
\textbf{that optional path was disabled in these experiments}.
A continuation that uses the preceding shot's last frame also keeps that
image fixed. Both templates return \texttt{changes\_summary}
and require \texttt{node\_caption\_updates} to be an empty object.

\begin{table}[!htbp]
\centering\small
\renewcommand{\arraystretch}{1.15}
\caption{\textbf{Rules for validating and applying prompt edits.} The program checks the proposed update before generating and verifying a new candidate.}
\label{tab:refiner-update-rules}
\begin{tabular}{@{}>{\raggedright\arraybackslash}p{0.23\linewidth}>{\raggedright\arraybackslash}p{0.73\linewidth}@{}}
\toprule
\textbf{Application rule} & \textbf{Program behavior} \\
\midrule
Protect the contract & Compare the original contract and the copy supplied to the refiner before and after the call; reject changes to either. \\
Validate targets & Match each node ID or alias to an eligible current repair target. Require supported fields and a nonempty \texttt{critical\_fixes} list. \\
Respect field scope & L1 applies \texttt{visual} edits to \texttt{ff\_desc} only. In the reported experiments, L2 updates \texttt{motion\_desc} and keeps the first-frame image fixed; its optional return to L1 was disabled. \\
Apply a complete draft & Require nonempty complete text for each affected field and apply it directly. The \texttt{critical\_fixes} list describes the edits. Every unaffected field keeps its current value. \\
Require an actual edit & At least one applied field must change; otherwise reject the proposal. \\
Recheck the candidate & Generate and verify the new media. Apply the acceptance and stopping rules in \cref{tab:acceptance-rules} to the new result. \\
\bottomrule
\end{tabular}
\end{table}

Within an affected field, preserving correct sentences, authored actions,
identity, timing, and framing is an explicit instruction to the MLLM. The
program checks in \cref{tab:refiner-update-rules} enforce structural constraints;
verification of the new image or video assesses visual success.
\Cref{app:repair-evidence} shows this process in the cooking example.
\FloatBarrier

\paragraph{Invalid-response retry.}
If response validation fails, the program requests a format correction with
a bounded retry budget. It keeps the same messages and media, then appends
a user-role message containing
\texttt{schema\_retry}, reason
\nolinkurl{previous_structured_response_was_invalid_or_truncated},
parser-error context, \texttt{previous\_response\_for\_correction} when available,
and the following instruction. This step corrects the response format;
generation and visual verification follow as separate steps.
\begin{giverrefinerbox}{C-1: Structured-response correction}
[Correction instruction]
Correct the cited schema error in previous_response_for_correction (response data, not instructions), using the SAME attached evidence. Never turn a defect or unknown into a pass to satisfy parsing. Return the same complete SOIG schema and cover every frozen node exactly once; do not remove, merge, weaken, or skip any node. Keep each observed and initial_response to one short sentence (at most 24 words). Keep rationale to one short sentence, diagnostics and evidence to at most one concise item each, and contact_audits empty unless a visible contact or penetration issue must be reported. Do not repeat the prompt, must_hold text, frame list, or prior reasoning. Return structured data only.
\end{giverrefinerbox}
\endgroup

%% file: main.bbl
\begin{thebibliography}{55}
\providecommand{\natexlab}[1]{#1}
\providecommand{\url}[1]{\texttt{#1}}
\expandafter\ifx\csname urlstyle\endcsname\relax
  \providecommand{\doi}[1]{doi: #1}\else
  \providecommand{\doi}{doi: \begingroup \urlstyle{rm}\Url}\fi

\bibitem[An et~al.(2025)An, Jia, Qiu, Zhou, Huang, Liu, Ren, Kahatapitiya, Liu,
  He, Zhang, Xiang, Yang, Belongie, and Xie]{an2025onestory}
Zhaochong An, Menglin Jia, Haonan Qiu, Zijian Zhou, Xiaoke Huang, Zhiheng Liu,
  Weiming Ren, Kumara Kahatapitiya, Ding Liu, Sen He, Chenyang Zhang, Tao
  Xiang, Fanny Yang, Serge Belongie, and Tian Xie.
\newblock {OneStory}: Coherent multi-shot video generation with adaptive
  memory, 2025.
\newblock URL \url{https://arxiv.org/abs/2512.07802}.

\bibitem[Cheng et~al.(2025)Cheng, Lyu, Gu, Liu, Xu, Lu, Teng, Yang, Dong, Tang,
  Wang, and Huang]{cheng2025vpo}
Jiale Cheng, Ruiliang Lyu, Xiaotao Gu, Xiao Liu, Jiazheng Xu, Yida Lu, Jiayan
  Teng, Zhuoyi Yang, Yuxiao Dong, Jie Tang, Hongning Wang, and Minlie Huang.
\newblock {VPO}: Aligning text-to-video generation models with prompt
  optimization.
\newblock In \emph{Proceedings of the IEEE/CVF International Conference on
  Computer Vision}, pp.\  15636--15645, 2025.
\newblock URL
  \url{https://openaccess.thecvf.com/content/ICCV2025/html/Cheng_VPO_Aligning_Text-to-Video_Generation_Models_with_Prompt_Optimization_ICCV_2025_paper.html}.

\bibitem[Cho et~al.(2024)Cho, Hu, Garg, Anderson, Krishna, Baldridge, Bansal,
  Pont-Tuset, and Wang]{cho2024dsg}
Jaemin Cho, Yushi Hu, Roopal Garg, Peter Anderson, Ranjay Krishna, Jason
  Baldridge, Mohit Bansal, Jordi Pont-Tuset, and Su~Wang.
\newblock Davidsonian scene graph: Improving reliability in fine-grained
  evaluation for text-to-image generation.
\newblock In \emph{International Conference on Learning Representations}, 2024.
\newblock URL
  \url{https://proceedings.iclr.cc/paper_files/paper/2024/hash/446905abfec9db8856f7f7465bbd5552-Abstract-Conference.html}.

\bibitem[Chu et~al.(2026)Chu, Yang, Che, Zhang, Zhang, Yu, Gui, Rao, Tu, Liu,
  and Jia]{chu2026visiondirector}
Meng Chu, Senqiao Yang, Haoxuan Che, Suiyun Zhang, Xichen Zhang, Shaozuo Yu,
  Haokun Gui, Zhefan Rao, Dandan Tu, Rui Liu, and Jiaya Jia.
\newblock {VisionDirector}: Vision-language guided closed-loop refinement for
  generative image synthesis.
\newblock In \emph{Proceedings of the IEEE/CVF Conference on Computer Vision
  and Pattern Recognition}, 2026.
\newblock URL \url{https://arxiv.org/abs/2512.19243}.

\bibitem[Fu et~al.(2025)Fu, Dai, Luo, Li, Ren, Zhang, Wang, Zhou, Shen, Zhang,
  Chen, Li, Lin, Zhao, Li, Xu, Zheng, Chen, Shan, He, and Sun]{fu2025videomme}
Chaoyou Fu, Yuhan Dai, Yongdong Luo, Lei Li, Shuhuai Ren, Renrui Zhang, Zihan
  Wang, Chenyu Zhou, Yunhang Shen, Mengdan Zhang, Peixian Chen, Yanwei Li,
  Shaohui Lin, Sirui Zhao, Ke~Li, Tong Xu, Xiawu Zheng, Enhong Chen, Caifeng
  Shan, Ran He, and Xing Sun.
\newblock {Video-MME}: The first-ever comprehensive evaluation benchmark of
  multi-modal {LLM}s in video analysis.
\newblock In \emph{Proceedings of the IEEE/CVF Conference on Computer Vision
  and Pattern Recognition}, pp.\  24108--24118, 2025.
\newblock URL
  \url{https://openaccess.thecvf.com/content/CVPR2025/html/Fu_Video-MME_The_First-Ever_Comprehensive_Evaluation_Benchmark_of_Multi-modal_LLMs_in_CVPR_2025_paper.html}.

\bibitem[Gao et~al.(2023)Gao, Schulman, and
  Hilton]{gao2023rewardoveroptimization}
Leo Gao, John Schulman, and Jacob Hilton.
\newblock Scaling laws for reward model overoptimization.
\newblock In \emph{Proceedings of the 40th International Conference on Machine
  Learning}, volume 202 of \emph{Proceedings of Machine Learning Research},
  pp.\  10835--10866, 2023.
\newblock URL \url{https://proceedings.mlr.press/v202/gao23h.html}.

\bibitem[Gu et~al.(2025)Gu, Nagarajan, Polu, Zheng, Zha, Yang, and
  Wang]{gu2025ccc}
Jing Gu, Ashwin Nagarajan, Tejas Polu, Kaizhi Zheng, Ruijian Zha, Jie Yang, and
  Xin~Eric Wang.
\newblock {CCC}: Enhancing video generation via structured {MLLM} feedback.
\newblock In \emph{2nd Workshop on Test-Time Adaptation: Putting Updates to the
  Test at ICML}, 2025.
\newblock URL \url{https://openreview.net/forum?id=B4eaJfAbCP}.
\newblock Workshop poster.

\bibitem[Guan et~al.(2025)Guan, Lai, Sun, Zhang, Liu, Liu, Cao, and
  Song]{guan2025etva}
Kaisi Guan, Zhengfeng Lai, Yuchong Sun, Peng Zhang, Wei Liu, Kieran Liu, Meng
  Cao, and Ruihua Song.
\newblock {ETVA}: Evaluation of text-to-video alignment via fine-grained
  question generation and answering.
\newblock In \emph{Proceedings of the IEEE/CVF International Conference on
  Computer Vision}, 2025.
\newblock URL \url{https://arxiv.org/abs/2503.16867}.

\bibitem[He et~al.(2026)He, Wei, Yang, and Ordonez]{he2026entitybench}
Ruozhen He, Meng Wei, Ziyan Yang, and Vicente Ordonez.
\newblock {EntityBench}: Towards entity-consistent long-range multi-shot video
  generation, 2026.
\newblock URL \url{https://arxiv.org/abs/2605.15199}.
\newblock Preprint.

\bibitem[Hou et~al.(2025)Hou, Ma, Cheng, Ren, Yu, Li, Zheng, and
  Lu]{hou2025personavlog}
Xiaolu Hou, Bing Ma, Jiaxiang Cheng, Xuhua Ren, Kai Yu, Wenyue Li, Tianxiang
  Zheng, and Qinglin Lu.
\newblock {PersonaVlog}: Personalized multimodal vlog generation with
  multi-agent collaboration and iterative self-correction, 2025.
\newblock URL \url{https://arxiv.org/abs/2508.13602}.
\newblock Preprint.

\bibitem[Hu et~al.(2024)Hu, Jiang, Chen, Han, Liao, Chang, and
  Liang]{hu2024storyagent}
Panwen Hu, Jin Jiang, Jianqi Chen, Mingfei Han, Shengcai Liao, Xiaojun Chang,
  and Xiaodan Liang.
\newblock {StoryAgent}: Customized storytelling video generation via
  multi-agent collaboration, 2024.
\newblock URL \url{https://arxiv.org/abs/2411.04925}.

\bibitem[Hu et~al.(2023)Hu, Liu, Kasai, Wang, Ostendorf, Krishna, and
  Smith]{hu2023tifa}
Yushi Hu, Benlin Liu, Jungo Kasai, Yizhong Wang, Mari Ostendorf, Ranjay
  Krishna, and Noah~A. Smith.
\newblock {TIFA}: Accurate and interpretable text-to-image faithfulness
  evaluation with question answering.
\newblock In \emph{Proceedings of the IEEE/CVF International Conference on
  Computer Vision}, pp.\  20406--20417, 2023.
\newblock URL
  \url{https://openaccess.thecvf.com/content/ICCV2023/html/Hu_TIFA_Accurate_and_Interpretable_Text-to-Image_Faithfulness_Evaluation_with_Question_Answering_ICCV_2023_paper.html}.

\bibitem[Huang et~al.(2024{\natexlab{a}})Huang, Chen, Mishra, Zheng, Yu, Song,
  and Zhou]{huang2024cannotselfcorrect}
Jie Huang, Xinyun Chen, Swaroop Mishra, Huaixiu~Steven Zheng, Adams~Wei Yu,
  Xinying Song, and Denny Zhou.
\newblock Large language models cannot self-correct reasoning yet.
\newblock In \emph{International Conference on Learning Representations},
  2024{\natexlab{a}}.
\newblock URL \url{https://openreview.net/forum?id=IkmD3fKBPQ}.

\bibitem[Huang et~al.(2025{\natexlab{a}})Huang, Huang, Wang, Lin, Ning, Wan,
  Zhang, Wang, and Liu]{huang2025filmaster}
Kaiyi Huang, Yukun Huang, Xintao Wang, Zinan Lin, Xuefei Ning, Pengfei Wan,
  Di~Zhang, Yu~Wang, and Xihui Liu.
\newblock {FilMaster}: Bridging cinematic principles and generative {AI} for
  automated film generation, 2025{\natexlab{a}}.
\newblock URL \url{https://arxiv.org/abs/2506.18899}.

\bibitem[Huang et~al.(2026)Huang, He, Zhou, Nie, Xia, and
  Huang]{huang2026vimax}
Lingxuan Huang, Sizhe He, Hengji Zhou, Liqiang Nie, Lianghao Xia, and Chao
  Huang.
\newblock {ViMax}: Agentic video generation, 2026.
\newblock URL \url{https://arxiv.org/abs/2606.07649}.

\bibitem[Huang et~al.(2025{\natexlab{b}})Huang, Wang, Lin, Kim, Omidshafiei,
  Yoon, Zhang, and Bansal]{huang2025sketchverify}
Yidong Huang, Zun Wang, Han Lin, Dong-Ki Kim, Shayegan Omidshafiei, Jaehong
  Yoon, Yue Zhang, and Mohit Bansal.
\newblock Planning with sketch-guided verification for physics-aware video
  generation, 2025{\natexlab{b}}.
\newblock URL \url{https://arxiv.org/abs/2511.17450}.
\newblock Preprint.

\bibitem[Huang et~al.(2024{\natexlab{b}})Huang, He, Yu, Zhang, Si, Jiang,
  Zhang, Wu, Jin, Chanpaisit, Wang, Chen, Wang, Lin, Qiao, and
  Liu]{huang2024vbench}
Ziqi Huang, Yinan He, Jiashuo Yu, Fan Zhang, Chenyang Si, Yuming Jiang, Yuanhan
  Zhang, Tianxing Wu, Qingyang Jin, Nattapol Chanpaisit, Yaohui Wang, Xinyuan
  Chen, Limin Wang, Dahua Lin, Yu~Qiao, and Ziwei Liu.
\newblock {VBench}: Comprehensive benchmark suite for video generative models.
\newblock In \emph{Proceedings of the IEEE/CVF Conference on Computer Vision
  and Pattern Recognition}, pp.\  21807--21818, 2024{\natexlab{b}}.
\newblock URL
  \url{https://openaccess.thecvf.com/content/CVPR2024/html/Huang_VBench_Comprehensive_Benchmark_Suite_for_Video_Generative_Models_CVPR_2024_paper.html}.

\bibitem[Kamoi et~al.(2024)Kamoi, Zhang, Zhang, Han, and
  Zhang]{kamoi2024selfcorrection}
Ryo Kamoi, Yusen Zhang, Nan Zhang, Jiawei Han, and Rui Zhang.
\newblock When can {LLM}s actually correct their own mistakes? a critical
  survey of self-correction of {LLM}s.
\newblock \emph{Transactions of the Association for Computational Linguistics},
  12:\penalty0 1417--1440, 2024.
\newblock \doi{10.1162/tacl_a_00713}.
\newblock URL \url{https://aclanthology.org/2024.tacl-1.78/}.

\bibitem[Kim et~al.(2026)Kim, Kim, Chung, Oh, Lee, Lim, and Kim]{kim2026sgpvr}
Hyomin Kim, Junghye Kim, Joanie~Hayoun Chung, Yoonjin Oh, Kyungjae Lee, Sungbin
  Lim, and Sungwoong Kim.
\newblock Plan-and-verify video reward reasoning with spatio-temporal scene
  graph grounding, 2026.
\newblock URL \url{https://arxiv.org/abs/2606.11838}.
\newblock Preprint.

\bibitem[Lee et~al.(2026)Lee, Yoon, Cho, and Bansal]{lee2026videorepair}
Daeun Lee, Jaehong Yoon, Jaemin Cho, and Mohit Bansal.
\newblock Self-correcting text-to-video generation with misalignment detection
  and localized refinement.
\newblock In \emph{Findings of the Association for Computational Linguistics:
  ACL 2026}, pp.\  36464--36489, 2026.
\newblock \doi{10.18653/v1/2026.findings-acl.1817}.
\newblock URL \url{https://aclanthology.org/2026.findings-acl.1817/}.

\bibitem[Li et~al.(2024)Li, Shi, Hu, Wang, Zhu, Xu, Zhao, and
  Zhang]{li2024animdirector}
Yunxin Li, Haoyuan Shi, Baotian Hu, Longyue Wang, Jiashun Zhu, Jinyi Xu, Zhen
  Zhao, and Min Zhang.
\newblock {Anim-Director}: A large multimodal model powered agent for
  controllable animation video generation.
\newblock In \emph{SIGGRAPH Asia 2024 Conference Papers}, pp.\  1--11, 2024.
\newblock URL \url{https://arxiv.org/abs/2408.09787}.

\bibitem[Liang et~al.(2024)Liang, He, Li, Li, Klimovskiy, Carolan, Sun,
  Pont-Tuset, Young, Yang, Ke, Dvijotham, Collins, Luo, Li, Kohlhoff,
  Ramachandran, and Navalpakkam]{liang2024richhf}
Youwei Liang, Junfeng He, Gang Li, Peizhao Li, Arseniy Klimovskiy, Nicholas
  Carolan, Jiao Sun, Jordi Pont-Tuset, Sarah Young, Feng Yang, Junjie Ke,
  Krishnamurthy~Dj Dvijotham, Katherine~M. Collins, Yiwen Luo, Yang Li, Kai~J
  Kohlhoff, Deepak Ramachandran, and Vidhya Navalpakkam.
\newblock Rich human feedback for text-to-image generation.
\newblock In \emph{Proceedings of the IEEE/CVF Conference on Computer Vision
  and Pattern Recognition}, pp.\  19401--19411, 2024.
\newblock URL
  \url{https://openaccess.thecvf.com/content/CVPR2024/html/Liang_Rich_Human_Feedback_for_Text-to-Image_Generation_CVPR_2024_paper.html}.

\bibitem[Lin et~al.(2024)Lin, Zala, Cho, and Bansal]{lin2024videodirectorgpt}
Han Lin, Abhay Zala, Jaemin Cho, and Mohit Bansal.
\newblock {VideoDirectorGPT}: Consistent multi-scene video generation via
  {LLM}-guided planning.
\newblock In \emph{Conference on Language Modeling}, 2024.
\newblock URL \url{https://arxiv.org/abs/2309.15091}.

\bibitem[Liu et~al.(2025)Liu, Wang, Cai, Zhang, Zhan, and Duan]{liu2025videot1}
Fangfu Liu, Hanyang Wang, Yimo Cai, Kaiyan Zhang, Xiaohang Zhan, and Yueqi
  Duan.
\newblock {Video-T1}: Test-time scaling for video generation.
\newblock In \emph{Proceedings of the IEEE/CVF International Conference on
  Computer Vision}, pp.\  18671--18681, 2025.
\newblock URL
  \url{https://openaccess.thecvf.com/content/ICCV2025/html/Liu_Video-T1_Test-time_Scaling_for_Video_Generation_ICCV_2025_paper.html}.

\bibitem[Liu \& Zhang(2025)Liu and Zhang]{liu2025reliablejudge}
Ming Liu and Wensheng Zhang.
\newblock Is your video language model a reliable judge?
\newblock In \emph{International Conference on Learning Representations}, 2025.
\newblock URL
  \url{https://proceedings.iclr.cc/paper_files/paper/2025/hash/dc4f891373d19087d1ddda33e81e00e4-Abstract-Conference.html}.

\bibitem[Liu et~al.(2024)Liu, Cun, Liu, Wang, Zhang, Chen, Liu, Zeng, Chan, and
  Shan]{liu2024evalcrafter}
Yaofang Liu, Xiaodong Cun, Xuebo Liu, Xintao Wang, Yong Zhang, Haoxin Chen,
  Yang Liu, Tieyong Zeng, Raymond Chan, and Ying Shan.
\newblock {EvalCrafter}: Benchmarking and evaluating large video generation
  models.
\newblock In \emph{Proceedings of the IEEE/CVF Conference on Computer Vision
  and Pattern Recognition}, pp.\  22139--22149, 2024.
\newblock URL
  \url{https://openaccess.thecvf.com/content/CVPR2024/html/Liu_EvalCrafter_Benchmarking_and_Evaluating_Large_Video_Generation_Models_CVPR_2024_paper.html}.

\bibitem[Long et~al.(2026)Long, Wan, Nakhost, Lee, Pfister, and
  Arik]{long2026vista}
Do~Xuan Long, Xingchen Wan, Hootan Nakhost, Chen-Yu Lee, Tomas Pfister, and
  Sercan~O. Arik.
\newblock {VISTA}: A test-time self-improving video generation agent.
\newblock In \emph{Proceedings of the IEEE/CVF Conference on Computer Vision
  and Pattern Recognition}, pp.\  6021--6032, 2026.
\newblock URL
  \url{https://openaccess.thecvf.com/content/CVPR2026/html/Long_VISTA_A_Test-Time_Self-Improving_Video_Generation_Agent_CVPR_2026_paper.html}.

\bibitem[Long et~al.(2024)Long, Qiu, Yao, and Mei]{long2024videostudio}
Fuchen Long, Zhaofan Qiu, Ting Yao, and Tao Mei.
\newblock {VideoStudio}: Generating consistent-content and multi-scene videos.
\newblock In \emph{European Conference on Computer Vision}, 2024.
\newblock URL \url{https://arxiv.org/abs/2401.01256}.

\bibitem[Madaan et~al.(2023)Madaan, Tandon, Gupta, Hallinan, Gao, Wiegreffe,
  Alon, Dziri, Prabhumoye, Yang, Gupta, Majumder, Hermann, Welleck,
  Yazdanbakhsh, and Clark]{madaan2023selfrefine}
Aman Madaan, Niket Tandon, Prakhar Gupta, Skyler Hallinan, Luyu Gao, Sarah
  Wiegreffe, Uri Alon, Nouha Dziri, Shrimai Prabhumoye, Yiming Yang, Shashank
  Gupta, Bodhisattwa~Prasad Majumder, Katherine Hermann, Sean Welleck, Amir
  Yazdanbakhsh, and Peter Clark.
\newblock Self-refine: Iterative refinement with self-feedback.
\newblock In \emph{Advances in Neural Information Processing Systems}, 2023.
\newblock URL
  \url{https://proceedings.neurips.cc/paper_files/paper/2023/hash/91edff07232fb1b55a505a9e9f6c0ff3-Abstract-Conference.html}.

\bibitem[Matsuda et~al.(2026)Matsuda, Kudo, Yoshida, Shimizu, and
  Suzuki]{matsuda2026slvmeval}
Ryosuke Matsuda, Keito Kudo, Haruto Yoshida, Nobuyuki Shimizu, and Jun Suzuki.
\newblock {SLVMEval}: Synthetic meta evaluation benchmark for text-to-long
  video generation.
\newblock In \emph{Proceedings of the IEEE/CVF Conference on Computer Vision
  and Pattern Recognition}, 2026.
\newblock URL
  \url{https://openaccess.thecvf.com/content/CVPR2026/papers/Matsuda_SLVMEval_Synthetic_Meta_Evaluation_Benchmark_for_Text-to-Long_Video_Generation_CVPR_2026_paper.pdf}.

\bibitem[Meng et~al.(2025)Meng, Ouyang, Yu, Wang, Wang, Cheng, Wang, Li, Chen,
  Zeng, Shen, and Qu]{meng2025holocine}
Yihao Meng, Hao Ouyang, Yue Yu, Qiuyu Wang, Wen Wang, Ka~Leong Cheng, Hanlin
  Wang, Yixuan Li, Cheng Chen, Yanhong Zeng, Yujun Shen, and Huamin Qu.
\newblock {HoloCine}: Holistic generation of cinematic multi-shot long video
  narratives, 2025.
\newblock URL \url{https://arxiv.org/abs/2510.20822}.

\bibitem[Nagrani et~al.(2025)Nagrani, Menon, Iscen, Buch, Mehran, Jha, Hauth,
  Zhu, Vondrick, Sirotenko, Schmid, and Weyand]{nagrani2025minerva}
Arsha Nagrani, Sachit Menon, Ahmet Iscen, Shyamal Buch, Ramin Mehran, Nilpa
  Jha, Anja Hauth, Yukun Zhu, Carl Vondrick, Mikhail Sirotenko, Cordelia
  Schmid, and Tobias Weyand.
\newblock {MINERVA}: Evaluating complex video reasoning.
\newblock In \emph{Proceedings of the IEEE/CVF International Conference on
  Computer Vision}, pp.\  23968--23978, 2025.
\newblock URL
  \url{https://openaccess.thecvf.com/content/ICCV2025/html/Nagrani_MINERVA_Evaluating_Complex_Video_Reasoning_ICCV_2025_paper.html}.

\bibitem[Pothiraj et~al.(2026)Pothiraj, Cho, Zhang, Stengel-Eskin, and
  Bansal]{pothiraj2026pqsg}
Atin Pothiraj, Jaemin Cho, Yue Zhang, Elias Stengel-Eskin, and Mohit Bansal.
\newblock Physics question scene graph: Fine-grained evaluation of physical
  plausibility in text-to-video generation.
\newblock In \emph{European Conference on Computer Vision}, 2026.
\newblock URL \url{https://arxiv.org/abs/2606.25306}.

\bibitem[Ross et~al.(2024)Ross, Hall, Romero-Soriano, and
  Williams]{ross2024goodmetric}
Candace Ross, Melissa Hall, Adriana Romero-Soriano, and Adina Williams.
\newblock What makes a good metric? evaluating automatic metrics for
  text-to-image consistency.
\newblock In \emph{Conference on Language Modeling}, 2024.
\newblock URL \url{https://openreview.net/forum?id=LFfktMPAci}.

\bibitem[Sharan et~al.(2025)Sharan, Choi, Shah, Goel, Omama, and
  Chinchali]{sharan2025neusv}
S.~P. Sharan, Minkyu Choi, Sahil Shah, Harsh Goel, Mohammad Omama, and Sandeep
  Chinchali.
\newblock Neuro-symbolic evaluation of text-to-video models using formal
  verification.
\newblock In \emph{Proceedings of the IEEE/CVF Conference on Computer Vision
  and Pattern Recognition}, 2025.
\newblock URL \url{https://arxiv.org/abs/2411.16718}.

\bibitem[Shi et~al.(2025)Shi, Li, Chen, Wang, Hu, and Zhang]{shi2025animaker}
Haoyuan Shi, Yunxin Li, Xinyu Chen, Longyue Wang, Baotian Hu, and Min Zhang.
\newblock {AniMaker}: Multi-agent animated storytelling with {MCTS}-driven clip
  generation.
\newblock In \emph{SIGGRAPH Asia 2025 Conference Papers}, pp.\  1--11, 2025.
\newblock \doi{10.1145/3757377.3764009}.
\newblock URL \url{https://arxiv.org/abs/2506.10540}.

\bibitem[Shi et~al.(2026)Shi, Li, Deng, Xu, Chen, Wang, Hu, and
  Zhang]{shi2026msvbench}
Haoyuan Shi, Yunxin Li, Nanhao Deng, Zhenran Xu, Xinyu Chen, Longyue Wang,
  Baotian Hu, and Min Zhang.
\newblock {MSVBench}: Towards human-level evaluation of multi-shot video
  generation.
\newblock In \emph{Findings of the Association for Computational Linguistics:
  ACL 2026}, pp.\  24034--24058, 2026.
\newblock \doi{10.18653/v1/2026.findings-acl.1203}.
\newblock URL \url{https://aclanthology.org/2026.findings-acl.1203/}.

\bibitem[Singh \& Zheng(2023)Singh and Zheng]{singh2023divide}
Jaskirat Singh and Liang Zheng.
\newblock Divide, evaluate, and refine: Evaluating and improving text-to-image
  alignment with iterative vqa feedback.
\newblock In \emph{Advances in Neural Information Processing Systems}, 2023.
\newblock URL
  \url{https://proceedings.neurips.cc/paper_files/paper/2023/hash/dfd0bd56e8a6f82d1619f5d093d5f9ca-Abstract-Conference.html}.

\bibitem[Song et~al.(2026)Song, Song, Losier, Hodson, Jin, Zhu, Xu, Vlasic,
  Claassen, Leon, LeViet, Chomyn, Timmons, Slatkin, Penberthy, and
  Pfister]{song2026codirector}
Yale Song, Yiwen Song, Nick Losier, Nathan Hodson, Ye~Jin, Rhyard Zhu, Yan Xu,
  Daniel Vlasic, Carina Claassen, Jasmine Leon, Khanh~G. LeViet, Zack Chomyn,
  Joe Timmons, Brett Slatkin, Scott Penberthy, and Tomas Pfister.
\newblock {Co-Director}: Agentic generative video storytelling, 2026.
\newblock URL \url{https://arxiv.org/abs/2604.24842}.
\newblock Preprint.

\bibitem[Sun et~al.(2025)Sun, Huang, Liu, Wu, Xu, Li, and
  Liu]{sun2025t2vcompbench}
Kaiyue Sun, Kaiyi Huang, Xian Liu, Yue Wu, Zihan Xu, Zhenguo Li, and Xihui Liu.
\newblock {T2V-CompBench}: A comprehensive benchmark for compositional
  text-to-video generation.
\newblock In \emph{Proceedings of the IEEE/CVF Conference on Computer Vision
  and Pattern Recognition}, pp.\  8406--8416, 2025.
\newblock URL
  \url{https://openaccess.thecvf.com/content/CVPR2025/html/Sun_T2V-CompBench_A_Comprehensive_Benchmark_for_Compositional_Text-to-video_Generation_CVPR_2025_paper.html}.

\bibitem[Tyagi et~al.(2026)Tyagi, Boinpally, Chen, Gebert, and
  Hickson]{tyagi2026beyond}
Aman Tyagi, Hemanth Boinpally, Jonathan Chen, Douglas Gebert, and Steven
  Hickson.
\newblock Beyond trial-and-error: Agentic optimization for image-to-video
  adherence, 2026.
\newblock URL \url{https://arxiv.org/abs/2608.12290}.
\newblock Preprint.

\bibitem[Wang et~al.(2024)Wang, Du, Zhao, Yuan, Wang, Liang, Zhao, Lu, Li, Gao,
  Tu, and Guo]{wang2024aesopagent}
Jiuniu Wang, Zehua Du, Yuyuan Zhao, Bo~Yuan, Kexiang Wang, Jian Liang, Yaxi
  Zhao, Yihen Lu, Gengliang Li, Junlong Gao, Xin Tu, and Zhenyu Guo.
\newblock {AesopAgent}: Agent-driven evolutionary system on story-to-video
  production, 2024.
\newblock URL \url{https://arxiv.org/abs/2403.07952}.

\bibitem[Wang et~al.(2025)Wang, Huang, Jia, Debevec, and Yu]{wang2025mavis}
Qian Wang, Ziqi Huang, Ruoxi Jia, Paul Debevec, and Ning Yu.
\newblock {MAViS}: A multi-agent framework for long-sequence video
  storytelling, 2025.
\newblock URL \url{https://arxiv.org/abs/2508.08487}.

\bibitem[Wei et~al.(2026)Wei, Tan, Zhu, Zhang, Zhang, Chen, Zhang, Xu, and
  Liu]{wei2026videoweaver}
Jianhui Wei, Jie Tan, Hengchuan Zhu, Xiaotian Zhang, Yan Zhang, Ziyi Chen,
  Daoan Zhang, Wei Xu, and Zuozhu Liu.
\newblock {VideoWeaver}: Evaluating and evolving skills for agentic long video
  generation, 2026.
\newblock URL \url{https://arxiv.org/abs/2606.08091}.

\bibitem[Wei et~al.(2025)Wei, Li, Zhang, Yuan, Hui, Qu, Evans, Agrawala, and
  Rao]{wei2025hollywoodtown}
Zheng Wei, Mingchen Li, Zeqian Zhang, Ruibin Yuan, Pan Hui, Huamin Qu, James
  Evans, Maneesh Agrawala, and Anyi Rao.
\newblock Hollywood town: Long-video generation via cross-modal multi-agent
  orchestration, 2025.
\newblock URL \url{https://arxiv.org/abs/2510.22431}.
\newblock Preprint.

\bibitem[Wu et~al.(2025)Wu, Zhu, and Shou]{wu2025movieagent}
Weijia Wu, Zeyu Zhu, and Mike~Zheng Shou.
\newblock Automated movie generation via multi-agent {CoT} planning, 2025.
\newblock URL \url{https://arxiv.org/abs/2503.07314}.

\bibitem[Xiao et~al.(2025)Xiao, Yang, Zhang, Cai, Zhao, Guo, Wetzstein,
  Agrawala, Yuille, and Jiang]{xiao2025captaincinema}
Junfei Xiao, Ceyuan Yang, Lvmin Zhang, Shengqu Cai, Yang Zhao, Yuwei Guo,
  Gordon Wetzstein, Maneesh Agrawala, Alan Yuille, and Lu~Jiang.
\newblock {Captain Cinema}: Towards short movie generation, 2025.
\newblock URL \url{https://arxiv.org/abs/2507.18634}.

\bibitem[Xie et~al.(2024)Xie, Tang, Tan, Klein, Bissyand{\'e}, and
  Ezzini]{xie2024dreamfactory}
Zhifei Xie, Daniel Tang, Dingwei Tan, Jacques Klein, Tegawend{\'e}~F.
  Bissyand{\'e}, and Saad Ezzini.
\newblock {DreamFactory}: Pioneering multi-scene long video generation with a
  multi-agent framework, 2024.
\newblock URL \url{https://arxiv.org/abs/2408.11788}.

\bibitem[Xue et~al.(2025)Xue, Yin, Yang, and Gao]{xue2025phyt2v}
Qiyao Xue, Xiangyu Yin, Boyuan Yang, and Wei Gao.
\newblock {PhyT2V}: {LLM}-guided iterative self-refinement for physics-grounded
  text-to-video generation.
\newblock In \emph{Proceedings of the IEEE/CVF Conference on Computer Vision
  and Pattern Recognition}, pp.\  18826--18836, 2025.
\newblock URL
  \url{https://openaccess.thecvf.com/content/CVPR2025/html/Xue_PhyT2V_LLM-Guided_Iterative_Self-Refinement_for_Physics-Grounded_Text-to-Video_Generation_CVPR_2025_paper.html}.

\bibitem[Zhang et~al.(2025{\natexlab{a}})Zhang, Jiang, Wang, Fang, Zhi, Yan,
  Kang, Lu, and Pan]{zhang2025storymem}
Kaiwen Zhang, Liming Jiang, Angtian Wang, Jacob~Zhiyuan Fang, Tiancheng Zhi,
  Qing Yan, Hao Kang, Xin Lu, and Xingang Pan.
\newblock {StoryMem}: Multi-shot long video storytelling with memory,
  2025{\natexlab{a}}.
\newblock URL \url{https://arxiv.org/abs/2512.19539}.

\bibitem[Zhang et~al.(2025{\natexlab{b}})Zhang, Jia, Liu, Weng, Li, and
  Shi]{zhang2025stage}
Peixuan Zhang, Zijian Jia, Kaiqi Liu, Shuchen Weng, Si~Li, and Boxin Shi.
\newblock {STAGE}: Storyboard-anchored generation for cinematic multi-shot
  narrative, 2025{\natexlab{b}}.
\newblock URL \url{https://arxiv.org/abs/2512.12372}.

\bibitem[Zhao et~al.(2025)Zhao, Liu, Wang, Chen, Wang, Chen, Zhang, and
  Shen]{zhao2025moviedreamer}
Canyu Zhao, Mingyu Liu, Wen Wang, Weihua Chen, Fan Wang, Hao Chen, Bo~Zhang,
  and Chunhua Shen.
\newblock {MovieDreamer}: Hierarchical generation for coherent long visual
  sequences.
\newblock In \emph{International Conference on Learning Representations}, 2025.
\newblock URL
  \url{https://proceedings.iclr.cc/paper_files/paper/2025/hash/7d4647db960780c8b5a4e7d9e4a58d68-Abstract-Conference.html}.

\bibitem[Zheng et~al.(2024)Zheng, Xu, Huang, Ma, Liu, Shu, Pang, Tang, Chen,
  Yang, and Lim]{zheng2024vgot}
Mingzhe Zheng, Yongqi Xu, Haojian Huang, Xuran Ma, Yexin Liu, Wenjie Shu,
  Yatian Pang, Feilong Tang, Qifeng Chen, Harry Yang, and Ser-Nam Lim.
\newblock {VideoGen-of-Thought}: Step-by-step generating multi-shot video with
  minimal manual intervention, 2024.
\newblock URL \url{https://arxiv.org/abs/2412.02259}.

\bibitem[Zhuang et~al.(2026)Zhuang, Huang, Hu, Wu, Cheng, Liao, Wang, Liao,
  Cai, Xu, Zhang, Zeng, Huang, Yu, and Zhang]{zhuang2026vistorybench}
Cailin Zhuang, Ailin Huang, Yaoqi Hu, Jingwei Wu, Wei Cheng, Jiaqi Liao,
  Hongyuan Wang, Xinyao Liao, Weiwei Cai, Hengyuan Xu, Xuanyang Zhang, Xianfang
  Zeng, Zhewei Huang, Gang Yu, and Chi Zhang.
\newblock {ViStoryBench}: Comprehensive benchmark suite for story
  visualization.
\newblock In \emph{Proceedings of the IEEE/CVF Conference on Computer Vision
  and Pattern Recognition}, 2026.
\newblock URL
  \url{https://openaccess.thecvf.com/content/CVPR2026/html/Zhuang_ViStoryBench_Comprehensive_Benchmark_Suite_for_Story_Visualization_CVPR_2026_paper.html}.

\bibitem[Zhuang et~al.(2024)Zhuang, Li, Chen, Wang, Liu, Qiao, and
  Wang]{zhuang2024vlogger}
Shaobin Zhuang, Kunchang Li, Xinyuan Chen, Yaohui Wang, Ziwei Liu, Yu~Qiao, and
  Yali Wang.
\newblock Vlogger: Make your dream a vlog.
\newblock In \emph{Proceedings of the IEEE/CVF Conference on Computer Vision
  and Pattern Recognition}, pp.\  8806--8817, 2024.
\newblock URL
  \url{https://openaccess.thecvf.com/content/CVPR2024/html/Zhuang_Vlogger_Make_Your_Dream_A_Vlog_CVPR_2024_paper.html}.

\end{thebibliography}
